\documentclass[journal]{IEEEtran}
\usepackage{amsmath,amsfonts}
\usepackage{algorithmic}
\usepackage{algorithm}
\usepackage{array}
\usepackage[caption=false,font=normalsize,labelfont=sf,textfont=sf]{subfig}
\usepackage[top=0.7in,bottom=0.5in,left=0.5in,textwidth=7.5in]{geometry}
\usepackage{textcomp}
\usepackage{stfloats}
\usepackage{url}
\usepackage{verbatim}
\usepackage{graphicx}
\usepackage{cite}
\usepackage{makecell} 
\usepackage[pagebackref=false,breaklinks=true,linkcolor=red,anchorcolor=blue, citecolor=green,colorlinks,bookmarks=false]{hyperref}
\begin{document}

\title{Cross Domain Object Detection via Multi-Granularity Confidence Alignment based Mean Teacher}

\author{Jiangming Chen, Li Liu, Wanxia Deng, Zhen Liu, Yu Liu, Yingmei Wei, Yongxiang Liu
\thanks{
This work was supported by the National Key Research and Development Program of China No. 2021YFB3100800, the National Natural Science Foundation of China under Grant 62376283, 62306323, and the Key Stone Grant JS2023-03 of the National University of Defense Technology. Li Liu is the corresponding author.}
\thanks{Jiangming Chen, Yu Liu, and Yingmei Wei are with the College of System Engineering, National University of Defense Technology (NUDT), Changsha, 410073, China (email: jiangming\_chen@126.com; jasonyuliu@nudt.edu.cn; weiyingmei@nudt.edu.cn).}
\thanks{Li Liu, Zhen Liu, and Yongxiang Liu are with the College of Electronic Science and Technology, NUDT, Changsha, 410073, China (email: liuli\_nudt@nudt.edu.cn; zhen\_liu@nudt.edu.cn; lyx\_bible@sina.com).}
\thanks{Wanxia Deng is with the College of Meteorology and Oceanography, NUDT, Changsha, 410073, China (email: wanxiadeng@163.com).}
}

\markboth{Submitted}
{Cheng \MakeLowercase{\textit{et al.}}:}


\maketitle

\begin{abstract}
Cross domain object detection learns an object detector for an unlabeled target domain by transferring knowledge from an annotated source domain. Promising results have been achieved via Mean Teacher, however, pseudo labeling which is the bottleneck of mutual learning remains to be further explored. In this study, we find that confidence misalignment of the predictions, including category-level overconfidence, instance-level task confidence inconsistency, and image-level confidence misfocusing, leading to the injection of noisy pseudo label in the training process, will bring suboptimal performance on the target domain. To tackle this issue, we present a novel general framework termed Multi-Granularity Confidence Alignment Mean Teacher (MGCAMT) for cross domain object detection, which alleviates confidence misalignment across \mbox{category-,} instance-, and image-levels simultaneously to obtain high quality pseudo supervision for better teacher-student learning. Specifically, to align confidence with accuracy at category level, we propose Classification Confidence Alignment (CCA) to model category uncertainty based on Evidential Deep Learning (EDL) and filter out the category incorrect labels via an uncertainty-aware selection strategy. Furthermore, to mitigate the instance-level misalignment between classification and localization, we design Task Confidence Alignment (TCA) to enhance the interaction between the two task branches and allow each classification feature to adaptively locate the optimal feature for the regression. Finally, we develop imagery Focusing Confidence Alignment (FCA) adopting another way of pseudo label learning, i.e., we use the original outputs from the Mean Teacher network for supervised learning without label assignment to concentrate on holistic information in the target domain image. When these three procedures are integrated into a single framework, they tend to benefit from each other to improve the final performance from a cooperative learning perspective. Extensive experiments on multiple widely-used scenarios demonstrate superiority over existing methods by a large margin. For example, we achieve 55.9\% mAP on Foggy Cityscapes and 44.8\% mAP on BDD100K, exceeding the previous state-of-the-art by 4.7\% and 4.6\% mAP, respectively.  
\end{abstract}


\section{Introduction}
\IEEEPARstart{D}{eep} neural networks has brought remarkable success in object detection~\cite{liu2020deep} which seeks to locate object instances from
categories of interest in images. Nevertheless, the impressive progress depends on fully supervised learning with a substantial amount of annotated data. The well-trained models under label supervision are prone to suffer severe degradation on other unseen domain, since they tend to bias towards the data distribution of the known domain~\cite{yosinski2014transferable}. This deficiency significantly hinders real-world applications, such as autonomous driving and video analysis. Meanwhile, large-scale labeled datasets are labor-intensive and time-consuming to obtain, making it impractical to improve the generalization ability of models through supervised learning in novel domains.

To address these limitations, cross domain object detection~\cite{oza2023unsupervised, saito2019strong, xu2020exploring, rezaeianaran2021seeking, cao2023contrastive, li2022scan, li2022sigma, liu2022decompose, he2023bidirectional, jiang2021decoupled} is proposed to adapt the detector from an annotated source domain to an unlabeled target domain. A variety of studies apply a common paradigm to conduct cross domain object detection, \emph{i.e.}, they introduce Unsupervised Domain Adaptation (UDA)~\cite{chen2022semi, deng2021joint} to align distributions of different domains. By this means, the detector can transfer the essential knowledge learned from the source domain to the target domain without additional manual annotations. In recent times, Mean Teacher (MT)~\cite{tarvainen2017mean} method shows great effectiveness in context information mining in the target domain and achieves promising results~\cite{li2022cross, deng2023harmonious}. The typical procedure is first to train a teacher detector using labeled source data, and then generate pseudo labels for the unlabeled target images, serving as the Ground Truth (GT) for the student detector. Subsequently, we optimize the student detector based on the consistency principle~\cite{tarvainen2017mean} which expects the model to produce concordant predictions regardless of data augmentation. Additionally, the teacher detector is a moving ensembling~\cite{laine2016temporal,li2022cross} of the student for high quality predictions. Mean Teacher leverages teacher-student mutual learning to improve the domain transferability of the detector progressively. 

\begin{figure}[t]
\centering
\centerline{\includegraphics[width=78mm]{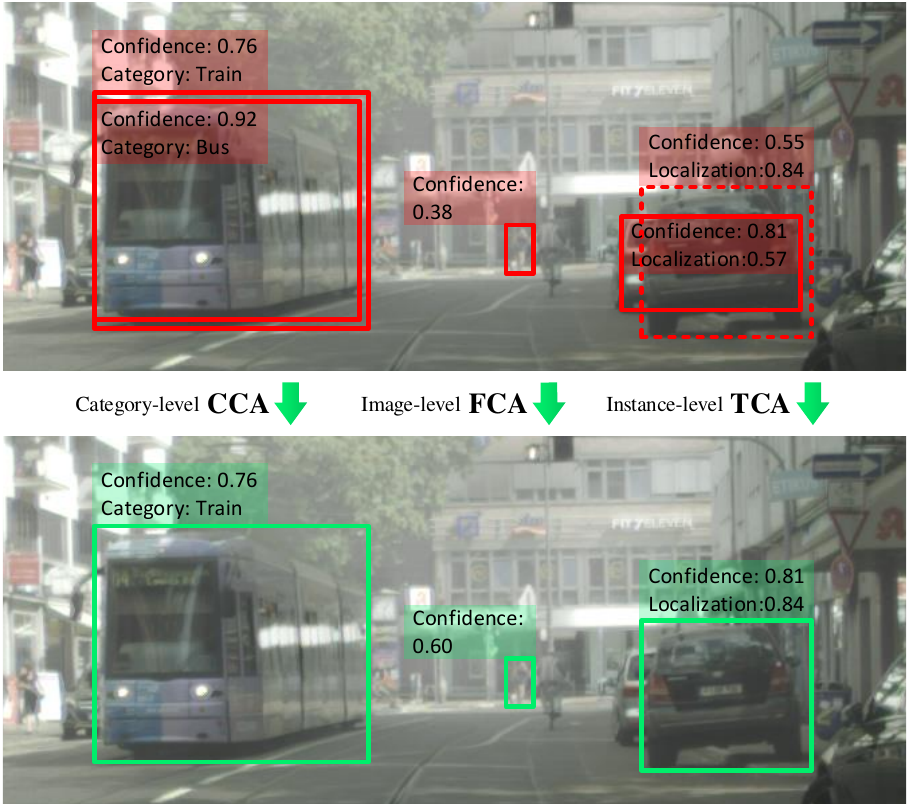}}
\caption{Illustration of our multi-granularity confidence alignment method for cross domain object detection. The predictions on the target domain (Foggy Cityscapes) are usually confidence misaligned: (left) the category with high confidence may be incorrect; (middle) the vague and tiny objects usually obtain low confidences or even get omitted; (right) the bounding box with high classification confidence is located inaccurately. From the perspective of alleviating confidence misalignment, we obtain high quality pseudo supervision by category-level CCA, instance-level TCA and image-level FCA to promote teacher-student mutual learning. (Best viewed in color and zooming in.)}
\label{fig:motivation}
\end{figure}

Despite the general efficacy of Mean Teacher method, its performance may still be constrained by pseudo labeling which is the critical subassembly. A uniform resolution is to receive the predictions with high classification confidence as pseudo labels.
Nevertheless, a high confidence does not always guarantee category accuracy especially in the case of domain shift, which is referred to as the issue of overconfidence~\cite{guo2017calibration, munir2023bridging} at category level (see the classification of train in Fig.~\ref{fig:motivation}). Furthermore, since the task branches of the detector are independently optimized without interaction, inconsistency between classification and localization confidences at instance level occurs when inferring the target domain images. As shown in Fig.~\ref{fig:motivation}, bounding box of a car with high category scores deviates from the ground truth localization. At last, in complex cross domain scenarios, such as when there are vague or multi-scale objects in an image, detectors often focus on the salient areas while predicting the inconspicuous objects with low scores or even missing them (see the detection of person in Fig.~\ref{fig:motivation}). Therefore, the aforementioned confidence misalignment will inject some noise into the pseudo labels, though a relatively high confidence threshold is set. The left pictures in Fig.~\ref{fig:confidence} and Fig.~\ref{fig:task} depict that many incorrectly categorized and mispositioned predictions are unfortunately received as pseudo labels, while lots of instances are missed. The noise in the labels will inevitably mislead the context information mining on the target domain resulting in suboptimal performance. Worse still, it may lead to negative feedback when we adopt anchor based dense detector (e.g., RetinaNet) as vanilla detector, where the detector model deteriorates in the target domain under pseudo supervision as shown by the blue curve in Fig.~\ref{fig:learning_iteration}.

Considering the above issue, we propose a novel general framework - MGCAMT, namely Multi-Granularity Confidence Alignment Mean Teacher, for cross domain object detection. From the perspective of alleviating confidence misalignment across category-, instance-, and image-levels simultaneously, we obtain high quality pseudo supervision to promote teacher-student mutual learning. Specifically, in order to align confidence with accuracy at category level, we propose Classification Confidence Alignment (CCA) to estimate the category uncertainty of the pseudo labels based on Evidential Deep Learning (EDL)~\cite{sensoy2018evidential}, and devise an uncertainty-aware selection strategy which filters out the potential incorrect labels. EDL is able to jointly formulate classification learning and uncertainty modeling, and compute uncertainty using single forward pass, thus naturally meeting the demands of online learning in the Mean Teacher paradigm. To mitigate the instance-level misalignment between classification and localization, we design a Task Confidence Alignment (TCA) module to enhance the interaction between the two branches and allow each classification feature to adaptively locate the optimal feature for the regression task. As for the imagery confidence misfocusing, we first analyze the transmission mechanism of the pseudo labels and observe that box label assignment magnifies the misfocusing problem on the target domain leading to negative feedback. This paper propose Focusing Confidence Alignment (FCA) to adopt another way of pseudo label learning, i.e., we use the original outputs from the Mean Teacher network for supervised learning without label assignment. By this means, we prevent foreground-background class imbalance, while preserving rich information from the teacher detector. Hence, the detector can focus on holistic information in the target image. These three confidence alignment methods respectively achieve correct classification, good instance localization, and balanced image-level perception. Accurate category attributes and comprehensive profile information of instance can positively reinforce each other. Moreover, high quality predictions of category and instance profile contribute to good imagery context perception, and vice versa. When we couple different confidence alignment procedures into a single framework, they manifest in a mutually beneficial manner to improve the final performance.

The main contributions of this study are as follows: 
\begin{itemize}
\item[$\bullet$]We propose a Multi-Granularity Confidence Alignment based Mean Teacher (MGCAMT) framework to explore cross domain object detection from a perspective of alleviating the overconfidence at category level, instance confidence mismatch between classification and localization, and imagery confidence misfocusing simultaneously.

\item[$\bullet$]We design Cassification Confidence Alignment (CCA) to align confidence with accuracy by introducing Evidential Deep Learning (EDL) to cross domain object detection. To our knowledge, EDL has not been used to reduce the effect of poor network calibration in the pseudo labeling process.

\item[$\bullet$]We develop a Task Confidence Alignment (TCA) component to regularize the consistency of the classification and localization confidences. TCA searches the holistic features for the regression that best matches the classification.

\item[$\bullet$]We observe that box label assignment magnifies the misfocusing problem on the target domain which may lead to negative feedback. We adopt Focusing Confidence Alignment (FCA) to mitigate this issue, while eliminating the tedious process of pseudo label assignment.

\item[$\bullet$]Extensive experiments of various domain shift scenarios are conducted to demonstrate that our method significantly improves the baselines and clearly outperforms the state-of-the-art by a large margin.
\end{itemize}

\section{Related Works}
\subsection{Object Detection}
Object detection aims to recognize and localize object instances of pre-defined classes in images. Some works follow two-stage methods~\cite{ren2015faster, cai2018cascade}, which predefine anchor boxes to generate coarse region proposals then refine these candidates to give the final results. Despite satisfactory accuracy, these approaches require high computational costs. Lots of studies focus on one-stage pipeline~\cite{liu2016ssd, lin2017focal, law2018cornernet, tian2019fcos} that directly predict targets regarding each pixel or its anchor as a reference. Anchor-based one-stage detector (e.g., RetinaNet~\cite{lin2017focal}) strikes the right balance between accuracy and efficiency. In recent times, methods based on set prediction~\cite{carion2020end, zhu2020deformable} have made object detection more concise, but typically come with high resource consumption. Moreover, the overconfidence in categories and the confidence mismatch between classification and localization~\cite{krishnan2020improving, munir2023bridging, feng2021tood,li2020generalized} hinder the potential exploration of the detector. To alleviate this issue, many studies~\cite{munir2023bridging, feng2021tood, li2020generalized, wang2021reconcile} have proposed a series of methods for classification-centric confidence alignment. Considering the great potential in real-world applications, we follow~\cite{pasqualino2021unsupervised, tian2023domain} to explore the cross domain transferability of the Anchor-based one-stage detector.

\subsection{Cross Domain Object Detection}
Cross domain object detection leverages a labeled source domain to learn an object detector which performs well in a novel unlabeled target domain. Many existing methods resort to matching distributions of the features between the source and target domains, such as adversarial training~\cite{saito2019strong, xu2020exploring, zhou2022multi}, contrastive learning~\cite{rezaeianaran2021seeking, vs2023towards, cao2023contrastive}, graph method~\cite{li2022scan, li2022sigma}, feature disentanglement~\cite{liu2022decompose, chen2023variational, he2023bidirectional}, etc. Recently, some studies~\cite{jiang2021decoupled, li2022cross, chen2022learning, deng2023harmonious} attempt to leverage the model itself to generate pseudo labels~\cite{lee2013pseudo} for unlabeled data in unsupervised domain adaptive learning. D\_adapt~\cite{jiang2021decoupled} utilizes pseudo labels in an offline manner. It decouples the adversarial adaptation from the detector optimization to maintain the discriminability of the model and alleviate the confirmation bias of pseudo labels. Mean Teacher (MT)~\cite{li2022cross, chen2022learning, deng2023harmonious} methods utilize pseudo labels online, and learn the context information of the target domain based on consistency regularization. With progressive learning and extensive data augmentation, it achieves promising results. AT~\cite{li2022cross} combines Mean Teacher with adversarial training to improve the quality of pseudo labels and stabilize pseudo label learning. PT~\cite{chen2022learning} represents the regression prediction as a probability distribution consistent with the form of classification and adjusts the weight of the pseudo label loss utilizing the entropy of the distribution. HT~\cite{deng2023harmonious} adopts Harmonious Loss to enforce the classification branch to predict a score in accord with the IoU between the predicted box and its GT, and weights the pseudo label loss with both the classification and localization scores. In this study, we propose a novel general framework termed Multi-Granularity Confidence Alignment Mean Teacher (MGCAMT) for cross domain object detection, which alleviates confidence misalignment across category-, instance-, and image-levels simultaneously to obtain high quality pseudo supervision for better teacher-student learning. At category level, we introduce Evidential Deep Learning (EDL)~\cite{sensoy2018evidential} to evaluate whether the category prediction is overconfident, and devise a selection strategy to filter out potential incorrect ones. At instance level, we allow each classification feature to adaptively locate the optimal feature for the regression through interactive remapping for better matches. At image level, we use the original outputs from the Mean Teacher network for supervised learning without label assignment to fix confidence misfocusing. We couple these three confidence alignment methods in a mutually beneficial manner to improve the performance. Our MGCAMT outperforms AT, PT, and HT separately by 5.0\%, 13.2\%, and 5.5\% on Foggy Cityscapes demonstrating our superiority.

\subsection{Evidential Deep Learning (EDL)}
Evidential deep learning (EDL)~\cite{sensoy2018evidential} is proposed to estimate uncertainty based on Evidence Theory~\cite{dempster2008generalization} and Subjective Logic~\cite{jsang2018subjective}, where the prediction of neural network is interpreted as a high order distribution on categorical probabilities placed a Beta or Dirichlet prior, rather than a point estimate. By performing a single forward pass, EDL obtains the predict uncertainty, greatly saving computational costs. The early work~\cite{sensoy2018evidential} proposes using EDL to reason reliable classification uncertainty, without loss of performance. The authors design a predictive distribution for classification by placing a Dirichlet distribution. Deep Evidential Regression~\cite{amini2020deep} introduces the evidential theory to regression tasks by placing evidential priors over the original Gaussian likelihood functions. Recently, MULE~\cite{zhao2023open} introduces evidential deep learning into the field of multi-label learning in the form of beta distribution. HUA~\cite{park2022active} adopts EDL to estimate uncertainty for active learning of object detection. A module termed Model Evidence Head which predicts the model evidence independently of the class confidence is also proposed to make EDL adaptable to object detection. In this paper, we employ EDL to perform uncertainty estimation in the Mean Teacher framework. Our main motivation lies in its advantages for reliable uncertainty estimation via evidence aggregation and the fact that the pseudo labels are usually overconfident which requires alignment. We propose an evidence learning method for sigmoid activation to guide uncertainty estimation in object detection.

\section{Proposed Methodology}
\begin{figure*}[t]
\centering
\centerline{\includegraphics[width=190mm]{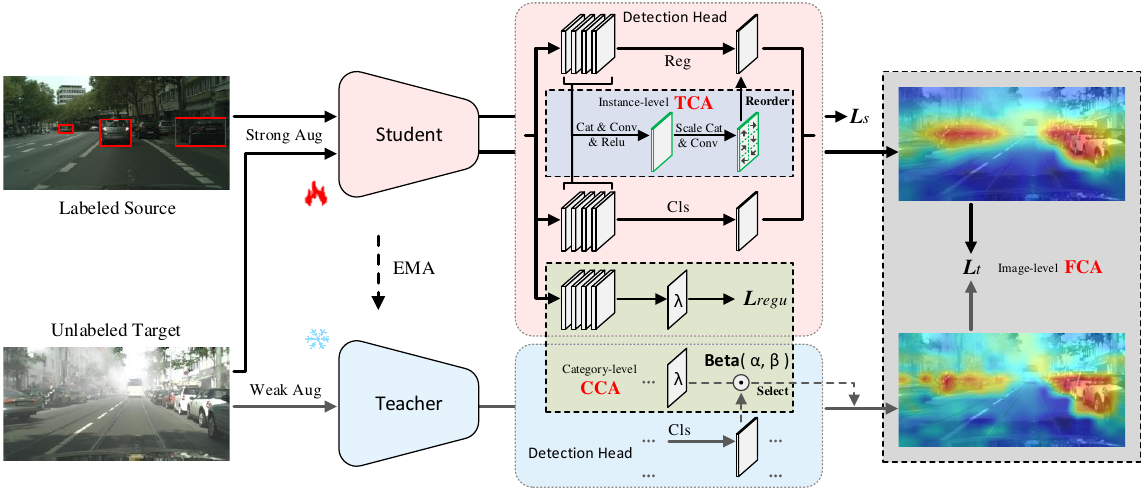}}
\caption{\textbf{Overview of the proposed Multi-Granularity Confidence Alignment Mean Teacher (MGCAMT)}. The student detector is optimized on the labeled source data
and the unlabeled target data with pseudo labels generated from the teacher detector. The student updates the teacher with exponential moving average (EMA). At category level, CCA adopts Beta-based evidential learning to estimate the category uncertainty of the pseudo labels, and reduces the effect of category overconfidence via an uncertainty-aware selection strategy. At instancel level, TCA enhances the interaction between the two task branches and allows each classification feature to adaptively locate the optimal feature for the regression based on remapping. At image level, FCA leverages the original outputs from the Mean Teacher network for supervised learning without label assignment to focus on holistic information of an image and eliminate the tedious process of pseudo label assignment. In the inference stage, only the convolution operation in TCA increases computational overhead $(\sim 1\%)$, but significantly improves the performance.}
\vspace{-0.2cm}
\label{fig:framework}
\end{figure*}

Formally, we are given $N_{s}$ labeled source domain samples $\mathcal{D}_{s} = \{(\textbf{x}_{i}^s, \textbf{y}_{i}^s)\}_{i = 1}^{N_{s}}$ and $N_{t}$ unlabeled target domain samples $\mathcal{D}_{t} = \{(\textbf{x}_{i}^t)\}_{i = 1}^{N_{t}}$, where $\textbf{x}$ denotes an image and $\textbf{y} = (\textbf{b}, \textbf{c})$ represents the annotations including the bounding box coordinates $\textbf{b}$ and the corresponding category labels $\textbf{c}$. The ultimate goal of cross domain object detection is to devise domain invariant detectors by leveraging $\mathcal{D}_{s}$ and $\mathcal{D}_{t}$. In the following, we introduce the proposed general framework named Multi-Granularity Confidence Alignment based Mean Teacher (MGCAMT) shown in Fig.~\ref{fig:framework}.

\subsection{Mean Teacher for Cross Domain Object Detection}
Following the Mean Teacher framework proposed for cross domain object detection in recent advances~\cite{li2022cross, chen2022learning, deng2023harmonious}, our model is composed of two architecturally identical submodels: a student detector and a teacher detector. The student detector is learned by standard gradient optimization, and the teacher detector is updated via a timely exponential moving average (EMA)~\cite{tarvainen2017mean} of the former. 

The unlabeled target domain images first go through weak augmentations and are fed into the teacher detector to generate pseudo labels, which are served as supervision signals for the corresponding images with strong augmentations. Subsequently, the student detector is trained on the labeled source domain data and the strongly augmented unlabeled target domain data with pseudo labels. Based on the consistency principle, the Mean Teacher framework fully mine the context information of the target domain images and effectively learn the invariant features. Hence, the overall training objective consists of a source domain loss $L_{s}$ and a target domain loss $L_{t}$:

\begin{equation}
\begin{aligned}
    L_{s} = L_{cls}(\textbf{x}^s, \textbf{y}^s) + L_{reg}(\textbf{x}^s, \textbf{y}^s),
    \label{l_s}
\end{aligned}
\end{equation}
\begin{equation}
\begin{aligned}
    L_{t} = L_{cls}(\textbf{x}^t, \textbf{y}^t) + L_{reg}(\textbf{x}^t, \textbf{y}^t),
    \label{l_t}
\end{aligned}
\end{equation}
where $L_{cls}$ is the classification loss, $L_{reg}$ is the box regression loss, $\textbf{y}^s$ is the Ground Truth (GT), and $\textbf{y}^t$ is the pseudo label annotated by the teacher detector. The total loss is as follows:
\begin{equation}
\begin{aligned}
    L_{total} = L_{s} + \omega L_{t}.
\label{l_total}
\end{aligned}
\end{equation}
where $\omega$ is a trade-off hyper-parameter. The EMA used for updating the teacher detector can be defined as: ${\theta_{t}} \leftarrow \delta\theta_{t} + (1 -\delta)\theta_{s}$, where $\theta_{t}$ and $\theta_{s}$ denote the network parameters of the teacher detector and student detector, respectively. We adopt RetinaNet~\cite{lin2017focal} as the base detector. Without specific notation, Focal Loss~\cite{lin2017focal} and GIoU loss~\cite{rezatofighi2019generalized} are set for $L_{cls}$ and $L_{reg}$ in this study.

\subsection{EDL Guided Classification Confidence Alignment}
\begin{figure}[t]
\centering
\centerline{\includegraphics[width=85mm]{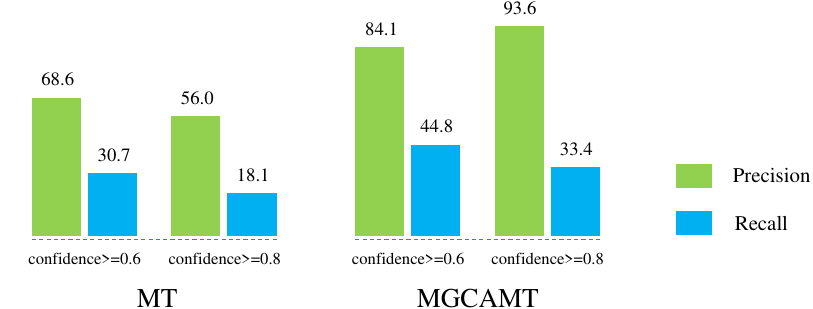}}
\caption{Classification precision of the predictions on the target domain when the confidence threshold is set to high levels. Meanwhile, we report the recall at the same confidence level. Results on the Foggy Cityscapes are presented. (Best viewed in color.)}
\label{fig:confidence}
\end{figure}

In common cross domain object detection frameworks, pseudo labels for the target domain images are generated purely according to category confidence. However, a high prediction confidence does not always guarantee accurate classification, which is referred to as the issue of overconfidence~\cite{guo2017calibration, munir2023bridging} at category level. As shown in the left of Fig.~\ref{fig:confidence}, the category precision does not improve with the increasing confidence, while the recall decreases dramatically. The poor alignment contributes to noise in pseudo labels misleading the context information mining on the target domain.

In order to align confidence with accuracy to promote the Mean Teacher network, we introduce Evidential Deep Learning (EDL)~\cite{sensoy2018evidential} to estimate the uncertainty of category as a reliability reference, and then devise a label selection strategy. We term the method Classification Confidence Alignment (CCA). EDL jointly formulates the classification learning and uncertainty modeling, and computes uncertainty in a single forward pass, thus naturally meeting the demands of online learning in the Mean Teacher paradigm. To the best of our knowledge, Evidential Deep Learning has not been used to reduce the effect of poor network calibration in the target domain pseudo labeling process.

1) Beta-based evidential learning for uncertainty estimation: EDL predicts credible uncertainty estimate by collecting evidence from each output class via introducing a high order distribution to represent the density of category probabilities $\textbf{p}$. Since RetinaNet empirically adopts Binomial likelihood to model multi-class classification problem as follows:
\begin{equation}
\begin{aligned}
    Binomial(y;p,1-p)=p^{y}(1-p)^{1-y},
    \label{Binomial}
\end{aligned}
\end{equation}
where $y \in \{0,1\}$, $p$ represent the label and corresponding category probability of a single class, we set the high order distribution as Beta distribution~\cite{zhao2023open} which is the conjugate prior of the likelihood. The Beta distribution with concentration parameters $\alpha, \beta > 0$ has a probability density:
\begin{equation}
\begin{aligned}
    Beta(p; \alpha, \beta) = \frac{1}{\mathrm{B}(\alpha, \beta)} p^{\alpha -1}(1-p)^{\beta -1},
    \label{beta_distribution}
\end{aligned}
\end{equation}
where $\mathrm{B}(\alpha, \beta) = \Gamma(\alpha)\Gamma(\beta) / \Gamma(\alpha+\beta)$, and $\Gamma(\cdot)$ indicates the gamma function. It is generalized from the Dirichlet distribution in EDL. 

\begin{figure}[t]
    \begin{minipage}[t]{0.5\linewidth}
        \centering
        \includegraphics[width=\textwidth]{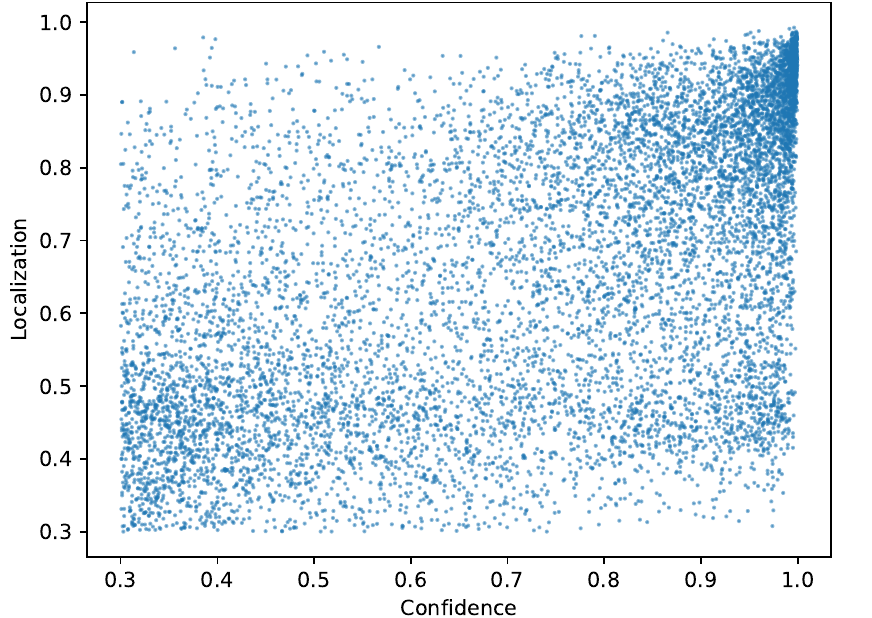}
        \centerline{\fontsize{8}{16} \selectfont MT}
    \end{minipage}%
    \begin{minipage}[t]{0.5\linewidth}
        \centering
        \includegraphics[width=\textwidth]{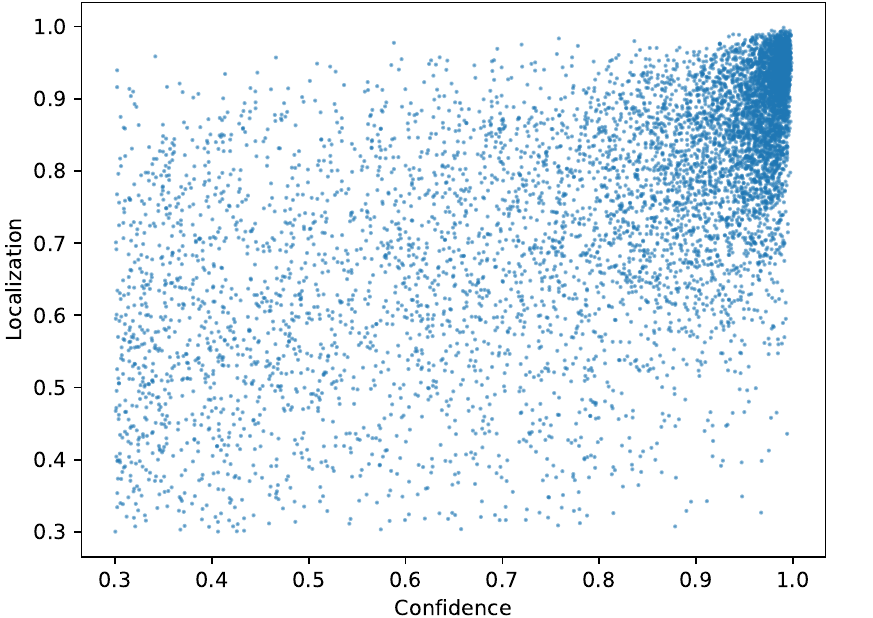}
        \centerline{\fontsize{8}{16} \selectfont MGCAMT}
    \end{minipage}
    \caption{Confidence of the maximum category and its maximum IoU with the Ground Truth boxes in the corresponding class. Results on the Foggy Cityscapes are presented. (Best viewed in color and zooming in.)}
\label{fig:task}
\end{figure}

Then we utilize the type-II maximum likelihood version of the loss described in the original work~\cite{sensoy2018evidential} to maximize the marginal likelihood. Therefore, the loss function in EDL is defined as:  
\begin{equation}
\begin{aligned}
    L_{edl} = -\mathrm{log} \left(\int p^{y}(1-p)^{1-y} \frac{1}{\mathrm{B}(\alpha, \beta)} p^{\alpha -1}(1-p)^{\beta -1} dp\right),
    \label{edl_origin}
\end{aligned}
\end{equation}
where $\alpha = e(f(\textbf{x}; \theta))$, $\beta = \alpha * exp^{-f(\textbf{x}; \theta)}$, and $e$ denotes evidence function (e.g., sigmoid activation) to keep $\alpha$ positive, $f(\textbf{x}; \theta)$ is a single channel output per anchor of the classification branch $f(\cdot)$ parametrized by $\theta$. Thanks to the Beta-Categorical conjugacy, the equation can be written in a closed-form:
\begin{equation}
\begin{aligned}
    L_{edl} = & -y \mathrm{log} \frac{\alpha}{\alpha + \beta} - (1-y)\mathrm{log} \frac{\beta}{\alpha + \beta}\\
            = & -(y \mathrm{log} p + (1-y)\mathrm{log} (1-p)).
    \label{edl}
\end{aligned}
\end{equation}
The loss in Eq.\ref{edl} can be treated as the binary cross-entropy loss of classification. We further replace it with focal loss to perform hard example mining. Then the uncertainty $u$ can be calculated directly using Eq.\ref{uncertainty}:
\begin{equation}
\begin{aligned}
    u = Var(P) = \frac{\alpha \beta}{(\alpha + \beta + 1)(\alpha + \beta)^2}.
    \label{uncertainty}
\end{aligned}
\end{equation}

We use sigmoid activation as the evidence function which enables confident prediction, but it forces the Beta strength $S = \alpha + \beta$ to be 1. This makes the entropy of Beta distribution unchangeable and bereaves the model of the ability to predict sufficient evidence. To tackle this issue, CCA solely predicts model evidence strength $\lambda$ to rescale the concentration parameters $\alpha$, $\beta$ following~\cite{park2022active}:
\begin{equation}
\begin{aligned}
    (\alpha, \beta) = \lambda \, (\alpha^{\prime}, \beta^{\prime}),
    \label{lambda}
\end{aligned}
\end{equation}
where $\alpha^{\prime}$ and $\beta^{\prime}$ are the origin concentration parameters. The evidence strength prediction follows the principle of class agnosticism. Furthermore, we adopt a regularization loss $L_{regu}$ following~\cite{park2022active} to penalize the model to better predict the model evidence dynamically:
\begin{equation}
\begin{aligned}
    L_{regu} = \sum_{j=1}^{M_a} \left(l_{j}^{cls} - \frac{1}{\lambda_{j}}\right)^2,
    \label{regularization}
\end{aligned}
\end{equation}
where $l_{j}^{cls}$ is the classification loss corresponding to the $j$-th anchor, and $M_{a}$ denotes the number of anchors for a single image. This loss encourages CCA to decrease model evidence $\lambda_j$ when the classification loss $l_{j}^{cls}$ is large; similarly, CCA is guided to increase $\lambda_j$ for small $l_{j}^{cls}$.

2) Pseudo label selection with awareness of uncertainty:
Although confidence based selection enhances the accuracy of pseudo labels, the issue of overconfidence renders this solution insufficient - pseudo labels with high confidence may be incorrect. We propose a pseudo label selection process with awareness of uncertainty to negate the effects of poor alignment: by utilizing both the confidence and uncertainty of the teacher prediction, a more reliable subset of pseudo labels are used to train the student detector. Let $g_j \in \{0,1\}$ be a binary variable representing whether the pseudo label is selected in a image, where $g_j = 1$ when the $j$-th pseudo label is selected and $g_j = 0$ when the label is not selected. We obtain $g_j$ as follows:
\begin{equation}
\begin{aligned}
    g_j = \text{1}[p_j>\tau_p]\,\text{1}[u_j<\tau_u],
    \label{selection}
\end{aligned}
\end{equation}
where $\text{1}[\cdot]$ is the indicator function, $p_j$, $u_j$ denote the maximum class confidence and corresponding uncertainty of the $j$-th pseudo label, and $\tau_p$, $\tau_u$ are the thresholds. This last term, involving $u_j$, ensures the teacher prediction is sufficiently certain to be selected.

\subsection{Interactive Remapping based Task Confidence Alignment}
The mismatch between classification and regression at instance level is another challenging problem in object detection~\cite{krishnan2020improving, munir2023bridging, feng2021tood,li2020generalized}, in addition to the category overconfidence. In fact, bounding boxes with high classification scores may deviate from the ground truth localization, and vice versa as ploted in Fig.~\ref{fig:task}. It hinders the essence of the Mean Teacher paradigm as we rely heavily on the prediction score to filter pseudo labels. When we use the labels with imprecise location information as supervision signals for the target domain, it leads to a sub-optimal performance inevitably.

Therefore, we explore how to improve the Mean Teacher framework by aligning the classification confidence and the bounding box localization for high quality predictions. Inspired by TOOD~\cite{feng2021tood}, we introduce a Task Confidence Alignment (TCA) module to enhance an interaction between the two tasks and allow each classification feature to adaptively locate the optimal feature for the regression task.

Given the multi-scale features $\textbf{F}$, where $F(m, n, o)$ represents the spatial location $(m, n)$ at the $o$-th scale level, we aim to construct a remapping function as follows: 
\begin{equation}
\begin{aligned}
    \textbf{F}^{\prime} \leftarrow r(\textbf{F}).
    \label{mapping}
\end{aligned}
\end{equation}
The remapping $r$ reorganizes the feature map to execute the regression task, therefore, $\textbf{F}^{\prime}$ better aligns with the classification features. Unlike the single-scale feature remapping approach discussed in~\cite{feng2021tood}, we extend the process to cross-scale feature space, acknowledging that the optimal features for classification and regression tasks may belong to different scales~\cite{liu2018path}. Meanwhile, the construction process of our mapping integrates classification and regression features, which facilitates the interaction between the two tasks.

The proposed task alignment is conducted by a sub-component in the detection head that estimates the spatial shift with multi-scale features for regression. As demonstrated in Fig.~\ref{fig:framework}, we employ one layer consisting of Conv 3×3 (Relu (Conv 1×1)) at different scale levels and calculate an shift vector $\textbf{s} = (s_0, s_1, s_2) \in R^{3}$ for each prediction. Cat and Scale Cat perform interaction between tasks and neighbouring scales respectively via concatenation operation. We first reorder $\textbf{F}$ in a single feature map space as follows: 
\begin{equation}
\begin{aligned}
    F^{\prime}(m, n, o) \leftarrow F(m + s_0, n + s_1, o),
    \label{off_single_scale}
\end{aligned}
\end{equation}
where the equation is realized by a bilinear interpolation. Then we realize feature shift across different scales for $\textbf{F}$ using the following equation:
\begin{equation}
\begin{aligned}
    F^{\prime}(m, n, o) \leftarrow F^{\prime}(m^{\prime}, n^{\prime}, o + s_2),
    \label{off_multi_scale}
\end{aligned}
\end{equation}
where $m^{\prime}$ and $n^{\prime}$ indicate the corresponding coordinates of $m$ and $n$ at different scale levels. The above formula is carried out by a resizing of $F^{\prime} (:, :, o + \lfloor s_2 \rfloor + 1)$ followed by a weighted average with $F^{\prime} (:, :, o + \lfloor s_2 \rfloor)$, where $\lfloor \cdot \rfloor$ denotes the floor operation. Notably, the convolutional layers in TCA increase computational overhead slightly $(\sim 1\%)$, but improve the performance significantly.

\begin{figure}[t]
\centering
\centerline{\includegraphics[width=88mm]{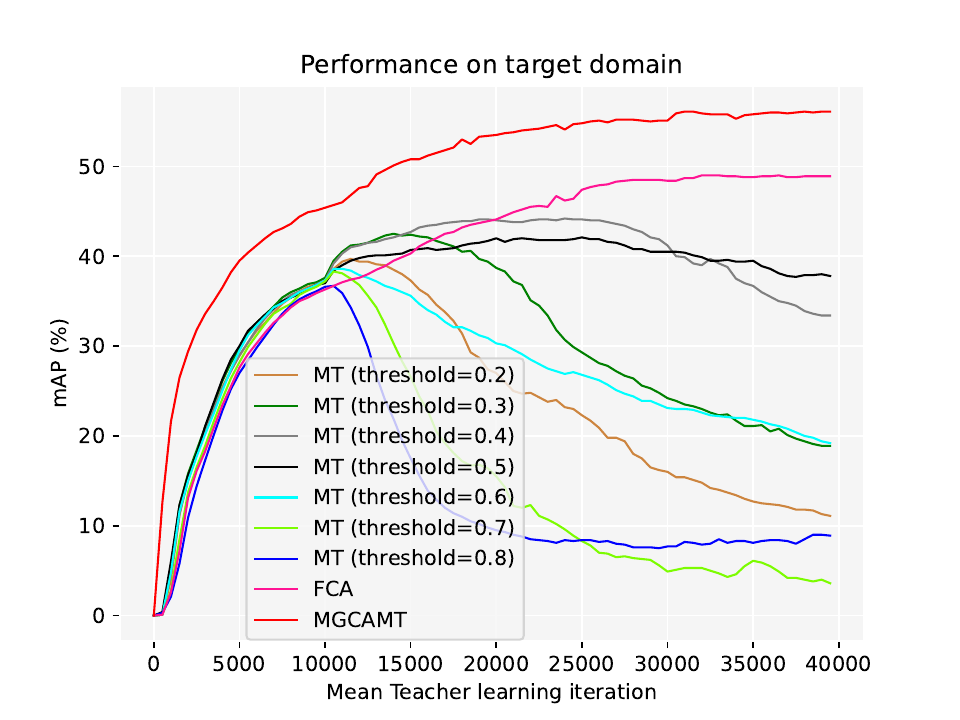}}
\caption{Illustration of negative feedback of pseudo label with label assignment. We adjust the confidence threshold to select pseudo label for Mean Teacher learning and we also report the performance of FCA and MGCAMT as a comparison. When the iteration reaches 10,000, pseudo signals participate in the training. Results on the Foggy Cityscapes are presented. (Best viewed in color.)}
\label{fig:learning_iteration}
\end{figure}

\subsection{Focusing Confidence Alignment without Label Assignment}
1) Negative feedback of pseudo label: The Mean Teacher paradigm typically generates pseudo labels as supervision signals for the target domain images, and then calculates the task losses after label assignment for cross domain learning. However, when this methodology comes to an anchor-based one-stage detector in complex cross domain scenarios, negative feedback emerges, where the detector deteriorates in the target domain under pseudo supervision shown by the blue curve in Fig.~\ref{fig:learning_iteration}. We illustrate some examples of predictions on Foggy Cityscapes ~\cite{sakaridis2018semantic} in the left of Fig.~\ref{fig:box} where we set the confidence $>=$ 0.6 for pseudo labeling. As can be seen, the pseudo labels annotate the salient areas well, but label the vague and tiny objects with low scores or even missing them due to the domain gap. We refer to this characteristic as confidence misfocusing.


In order to guarantee the completeness of pseudo labels as much as possible, we lower the confidence threshold, but this results in lots of redundant boxes which are predominantly distributed over salient areas (see the examples in the second column of Fig.~\ref{fig:box}). As shown in Fig.~\ref{fig:learning_iteration}, negative feedback still exists despite the threshold varies over a wide range. When the threshold is set to 0.5, the negative feedback problem is largely alleviated, but unfortunately, pseudo label learning has very limited effect on the enhancement of cross domain performance. Furthermore, we replace the pseudo labels with the Ground Truth labels to train the Mean Teacher model. With assistance of the GT, the detector significantly improves on the target domain.

Let’s revisit the transmission mechanism of labels in RetinaNet, where anchors are assigned to annotated object boxes using an intersection-over-union (IoU) threshold of 0.5; and to background if their IoU is in [0, 0.4). When some dense small objects are not annotated by pseudo labels, anchors tiled in this areas will be assigned as background samples. Actually, RetinaNet tiles 9 anchors (3 scales × 3 aspect ratios) per location. It leads to extreme foreground-background class imbalance. As a consequence, the detector predicts fewer and fewer foreground objects gradually. When there is redundancy in the pseudo labels, the detector will excessively focus on the redundant areas and ignore some correctly labeled regions, which also leads to negative feedback. 

2) Pseudo label learning without label assignment: To mitigate the negative feedback issue caused by the confidence misfocusing, this paper adopts another way of pseudo label learning, i.e., we use the original outputs from the Mean Teacher network for supervised learning without label assignment. By this means, we prevent foreground-background class imbalance, while preserving rich information from the teacher detector to explore content in areas of the target domain that are not significant. Notably, we discard Non-Maximum Suppression (NMS) as it may exacerbate the unreliability of pseudo labels~\cite{zhou2022dense}, especially in cases of category overconfidence and task inconsistency. This approach is named Focusing Confidence Alignment (FCA). Specifically, we utilize the post-sigmoid logits predicted by the teacher network as the desired pseudo labels. As the pseudo supervised signal obtained through activation is continuous, we substitute Quality Focal Loss~\cite{li2020generalized} for Focal Loss as follows:
\begin{equation}
\begin{aligned}
    l^{t}_{cls}(p,y) = & -\left|y - p \right|^{\gamma} \, (y \mathrm{log} p + (1-y) \mathrm{log} (1-p)).
    \label{l_cls_t}
\end{aligned}
\end{equation}

Then we have the following classification loss of target domain:
\begin{equation}
\begin{aligned}
    \hat{L}^{t}_{cls}(\textbf{x}^t, \textbf{y}^t) = \sum_{j=1}^{M_{f}}\sum_{k=1}^{K} l^{t}_{cls}(p_{j,k},y_{j,k}),
    \label{L_cls_t}
\end{aligned}
\end{equation}
where $y_{j,k}$ denotes the $k$-th output channel for the $j$-th anchor by the teacher detector and $p_{j,k}$ is the corresponding output of the student detector. $M_{f}$ stands for the number of filtered anchors for a single image. $K$ represents the number of categories. It is worth noting that~\cite{zhou2022dense} proposes Dense Pseudo Label (DPL) for eliminating the tedious post-processing of pseudo label. Our work has a similar motivation to~\cite{zhou2022dense} but for different tasks and goals.

\subsection{Optimization and Inference}
By combining the above-mentioned constraints, the final objective of the proposed framework can be formulated as:  
\begin{equation}
\begin{aligned}
    L_{total} = L_{s} + \omega \hat{L}_{t} + L_{regu},
\label{L_total}
\end{aligned}
\end{equation}
where $L_s$ is the source domain loss, $\hat{L}_{t}$ is the target domain loss with classification loss replaced by $\hat{L}^{t}_{cls}$ in Eq.\ref{L_cls_t}, $L_{regu}$ is the evidence regularization loss, and $\omega$ is a trade-off hyper-parameter. We optimize it to adapt the detector to the target domain.

During inference, we implement a vanilla object detector solely containing the TCA module to perform object detection, as the model has been well optimized for the cross domain scenario.

\begin{table*}[t]
\centering
\caption{Experimental results from Cityscapes to Foggy Cityscapes. We use VGG16 as the backbone network. Average precision (\%) is evaluated on target images.} \label{tab:ctf}
\renewcommand{\arraystretch}{1}
\setlength\tabcolsep{2pt}
\setlength\arrayrulewidth{0.2mm}
\begin{tabular}{c|c|c|*{8}{>{\centering\arraybackslash}p{0.8cm}}|*{1}{>{\centering\arraybackslash}p{0.8cm}}}
  \hline
  Method & Venue & Detector & person & rider & car & truck & bus & train & motor & bicycle & mAP
  \\
  \hline
  SWDA~\cite{saito2019strong} & CVPR 2019 & Faster R-CNN & 29.9 & 42.3 & 43.5 & 24.5 & 36.2 & 32.6 & 30.0 & 35.3 & 34.3\\
  HTCN~\cite{chen2020harmonizing} & CVPR 2020 & Faster R-CNN & 33.2 & 47.5 & 47.9 & 31.6 & 47.4 & 40.9 & 32.3 & 37.1 & 39.8 \\
  SIR~\cite{chen2021sequential} & TIP 2021 & Faster R-CNN & 33.0 & 45.6 & 52.3 & 30.5 & 50.8 & 40.9 & 32.6 & 35.9 & 40.2\\
  D\_adapt~\cite{jiang2021decoupled} & ICLR 2022 & Faster R-CNN & 44.9 & 54.2 & 61.7 & 25.6 & 36.3 & 24.7 & 37.3 & 46.1 & 41.3\\
  TDD\cite{he2022cross} & CVPR 2022 & Faster R-CNN & 39.6 & 47.5 & 55.7 & 33.8 & 47.6 & 42.1 & 37.0 & 41.4 & 43.1 \\
  AT\cite{li2022cross} & CVPR 2022 & Faster R-CNN & 45.5 &55.1 & 64.2 & 35.0 & 56.3 & 54.3 & 38.5 & 51.9 & 50.9 \\
  PT\cite{chen2022learning} & ICML 2022 & Faster R-CNN & 40.2 & 48.8 & 59.7 & 30.7 & 51.8 & 30.6 & 35.4 & 44.5 & 42.7 \\
  CMT~\cite{cao2023contrastive} & CVPR 2023 & Faster R-CNN & 45.9 & 55.7 & 63.7 & \textbf{39.6} & \textbf{66.0} & 38.8 & 41.4 & 51.2 & 50.3\\
  \hline
  EPM~\cite{hsu2020every} & ECCV 2020 & FCOS & 41.9 & 38.7 & 56.7 & 22.6 & 41.5 & 26.8 & 24.6 & 35.5 & 36.0\\
  SSAL~\cite{munir2021ssal} & NeurIPS 2021 & FCOS & 45.1 & 47.4 & 59.4 & 24.5 & 50.0 & 25.7 & 26.0 & 38.7 & 39.6\\
  SIGMA~\cite{li2022sigma} & CVPR 2022 & FCOS & 44.0 & 43.9 & 60.3 & 31.6 & 50.4 & 51.5 & 31.7 & 40.6 & 44.2\\
  OADA~\cite{yoo2022unsupervised} & ECCV 2022 & FCOS & 47.8 & 46.5 & 62.9 & 32.1 & 48.5 & 50.9 & 34.3 & 39.8 & 45.4\\
  CIGAR~\cite{liu2023cigar} & CVPR 2023 & FCOS & 46.1 & 47.3 & 62.1 & 27.8 & 56.6 & 44.3 & 33.7 & 41.3 & 44.9\\
  HT~\cite{deng2023harmonious} & CVPR 2023 & FCOS & 52.1 & 55.8 & 67.5 & 32.7 & 55.9 & 49.1 & 40.1 & 50.3 & 50.4\\
  \hline
  SFA~\cite{wang2021exploring} & ACM MM 2021 & Def DETR & 46.5 & 48.6 & 62.6 & 25.1 & 46.2 & 29.4 & 28.3 & 44.0 & 41.3\\
  AQT~\cite{huang2022aqt} & IJCAI 2022 & Def DETR & 49.3 & 52.3 & 64.4 & 27.7 & 53.7 & 46.5 & 36.0 & 46.4 & 47.1\\
  O$^2$Net~\cite{gong2022improving} & ACM MM 2022 & Def DETR & 48.7 & 51.5 & 63.6 & 31.1 & 47.6 & 47.8 & 38.0 & 45.9 & 46.8\\
  DA-DETR~\cite{zhang2023detr} & CVPR 2023 & Def DETR & 49.9 & 50.0 & 63.1 & 24.0 & 45.8 & 37.5 & 31.6 & 46.3 & 43.5\\
  BiADT~\cite{he2023bidirectional} & ICCV 2023 & Def DETR & 52.2 & 58.9 & 69.2 & 31.7 & 55.0 & 45.1 & 42.6 & 51.3 & 50.8\\
  MRT~\cite{zhao2023masked} & ICCV 2023 & Def DETR & 52.8 & 51.7 & 68.7 & 35.9 & 58.1 & 54.5 & 41.0 & 47.1 & 51.2\\
  MTM~\cite{weng2023mean} & AAAI 2024 & Def DETR & 51.0 & 53.4 & 67.2 & 37.2 & 54.4 & 41.6 & 38.4 & 47.7 & 48.9\\
  \hline
  Source Only & - & RetinaNet & 46.2 & 52.3 & 61.3 & 14.8 & 36.2 & 4.9 & 34.1 & 49.2 & 37.4\\
  DA-RetinaNet~\cite{pasqualino2021unsupervised} & CVIU 2021 & RetinaNet & 51.9 & 51.7 & 59.1 & 34.4 & 47.2 & 30.4 & 38.7 & 51.0 & 45.6\\
  KTNet~\cite{tian2023domain} & Neural Networks 2023 & RetinaNet & 43.7 & 48.9 & 56.9 & 34.9 & 51.8 & \textbf{55.3} & 32.8 & 42.8 & 45.9\\
  Ours & - & RetinaNet & \textbf{60.2} & \textbf{66.6} & \textbf{76.5} & 33.2 & 60.1 & 43.2 & \textbf{49.8} & \textbf{57.9} & \textbf{55.9}\\
  \hline
\end{tabular}
\end{table*}

\begin{table*}[t]
\centering
\caption{Experimental results from Cityscapes to BDD100K. We use VGG16 as the backbone network. Average precision (\%) is evaluated on target images.} \label{tab:ctb}
\renewcommand{\arraystretch}{1}
\setlength\tabcolsep{2pt}
\setlength\arrayrulewidth{0.2mm}
\begin{tabular}{c|c|c|*{7}{>{\centering\arraybackslash}p{0.8cm}}|*{1}{>{\centering\arraybackslash}p{0.8cm}}}
  \hline
  Method & Venue & Detector & person & rider & car & truck & bus & motor & bicycle & mAP
  \\
  \hline
  SWDA~\cite{saito2019strong} & CVPR 2019 & Faster R-CNN & 29.5 & 29.9 & 44.8 & 20.2 & 20.7 & 15.2 & 23.1 & 26.2\\
  ECR~\cite{xu2020exploring} & CVPR 2020 & Faster R-CNN & 32.8 & 29.3 & 45.8 & 22.7 & 20.6 & 14.9 & 25.5 & 27.4 \\
  SED~\cite{li2021free} & AAAI 2021 & Faster R-CNN & 32.4 & 32.6 & 50.4 & 20.6 & 23.4 & 18.9 & 25.0 & 29.0\\
  TDD\cite{he2022cross} & CVPR 2022 & Faster R-CNN & 39.6 & 38.9 & 53.9 & 24.1 & 25.5 & 24.5 & 28.8 & 33.6 \\
  PT\cite{chen2022learning} & ICML 2022 & Faster R-CNN & 40.5 & 39.9 & 52.7 & 25.8 & \textbf{33.8} & 23.0 & 28.8 & 34.9 \\
  \hline
  EPM~\cite{hsu2020every} & ECCV 2020 & FCOS & 39.6 & 26.8 & 55.8 & 18.8 & 19.1 & 14.5 & 20.1 & 27.8\\
  SIGMA~\cite{li2022sigma} & CVPR 2022 & FCOS & 46.9 & 29.6 & 64.1 & 20.2 & 23.6 & 17.9 & 26.3 & 32.7\\
  HT~\cite{deng2023harmonious} & CVPR 2023 & FCOS & 53.4 & 40.4 & 63.5 & 27.4 & 30.6 & 28.2 & \textbf{38.0} & 40.2\\
  \hline
  SFA~\cite{wang2021exploring} & ACM MM 2021 & Def DETR & 40.2 & 27.6 & 57.5 & 19.1 & 23.4 & 15.4 & 19.2 & 28.9\\
  AQT~\cite{huang2022aqt} & IJCAI 2022 & Def DETR & 38.2 & 33.0 & 58.4 & 17.3 & 18.4 & 16.9 & 23.5 & 29.4\\
  O$^2$Net~\cite{gong2022improving} & ACM MM 2022 & Def DETR & 40.4 & 31.2 & 58.6 & 20.4 & 25.0 & 14.9 & 22.7 & 30.5 \\
  BiADT~\cite{he2023bidirectional} & ICCV 2023 & Def DETR & 42.1 & 34.0 & 60.9 & 17.4 & 19.5 & 18.2 & 25.7 & 33.6\\
  MRT~\cite{zhao2023masked} & ICCV 2023 & Def DETR & 48.4 & 30.9 & 63.7 & 24.7 & 25.5 & 20.2 & 22.6 & 33.7\\
  MTM~\cite{weng2023mean} & AAAI 2024 & Def DETR & 53.7 & 35.1 & 68.8 & 23.0 & 28.8 & 23.8 & 28.0 & 37.3\\
  \hline
  Source Only & - & RetinaNet & 46.6 & 27.9 & 63.3 & 7.4 & 8.5 & 10.5 & 23.7 & 26.9\\
  DA-RetinaNet~\cite{pasqualino2021unsupervised} & CVIU 2021 & RetinaNet & 48.4 & 39.6 & 63.4 & 26.4 & 25.5 & 29.3 & 30.0 & 37.5\\
  Ours & - & RetinaNet & \textbf{61.2} & \textbf{44.9} & \textbf{75.3} & \textbf{32.5} & 30.0 & \textbf{31.9} & 37.8 & \textbf{44.8}\\
  \hline
\end{tabular}
\end{table*}

\section{Experiments}
\subsection{Dataset}
To evaluate the effectiveness of our method, extensive experiments are conducted on five public datasets following the standard cross domain object detection setting in previous works~\cite{hsu2020every, chen2022learning, he2023bidirectional} including Cityscapes, Foggy Cityscapes, BDD100K, KITTI, and Sim10K.

\textbf{Cityscapes}~\cite{cordts2016cityscapes} is a benchmark dataset for instance segmentation, which is captured with an on-board camera in urban street scenes. It contains 2,975 images in the training set and 500 images in the validation set, covering 8 object categories. We generate the bounding box annotations using the tightest rectangles of each instance mask. 

\textbf{Foggy Cityscapes}~\cite{sakaridis2018semantic}  simulates fog in real scenes with depth information by rendering images from Cityscapes. It has the same categories and number of images as CityScapes. The worst foggy level (i.e., 0.02) from the dataset is chosen for the experiments. This dataset aggregates a large number of dense small targets, while also suffering from severe foggy shift, presenting a significant challenge for us.

\textbf{BDD100K}~\cite{yu2020bdd100k} is a similar scene dataset to Cityscapes except that BDD100K has different camera setup, sharing 7 categories with Cityscapes. The subset of images labeled as daytime are used as our target domain, including 36,278 training and 5,258 validation images.

\textbf{KITTI}~\cite{geiger2012we} contains 7,481 labeled images with the car category for autonomous driving. The KITTI dataset and Cityscapes dataset have similar scenes on roads, but they are captured by different cameras. On the other hand, KITTI is larger than Cityscapes, and we use it for experiments on transferring between datasets of different sizes.

\textbf{Sim10K}~\cite{johnson2017driving} is a synthesized dataset, which is rendered from the Grand Theft Auto engine. It has 10,000 images annotated with only car category. Synthetic datasets play a crucial role as they do not require time-consuming and labor-intensive expert annotations, but can be derived solely from existing algorithms.

\begin{table}[t]
\centering
\caption{Experimental results from  KITTI to Cityscapes. We use VGG16 as the backbone network. Average precision (\%) is evaluated on target images.} \label{tab:ktc}
\renewcommand{\arraystretch}{1}
\setlength\tabcolsep{2pt}
\setlength\arrayrulewidth{0.2mm}
\begin{tabular}{c|c|c|c}
  \hline
  Method & Venue & Detector & AP of car\\
  \hline
  SCDA~\cite{zhu2019adapting} & CVPR 2019 & Faster R-CNN & 42.5\\
  HTCN~\cite{chen2020harmonizing} & CVPR 2020 & Faster R-CNN & 42.1 \\
  SIR~\cite{chen2021sequential} & TIP 2021 & Faster R-CNN & 46.5 \\
  SED~\cite{li2021free} & CVPR 2021 & Faster R-CNN & 43.7\\
  TDD\cite{he2022cross} & CVPR 2022 & Faster R-CNN & 47.4 \\
  PT\cite{chen2022learning} & ICML 2022 & Faster R-CNN & 60.2 \\
  \hline
  EPM~\cite{hsu2020every} & ECCV 2020 & FCOS & 43.2\\
  SSAL~\cite{munir2021ssal} & NeurIPS 2021 & FCOS & 45.6\\
  SIGMA~\cite{li2022sigma} & CVPR 2022 & FCOS & 45.8\\
  OADA~\cite{yoo2022unsupervised} & ECCV 2022 & FCOS & 47.8\\
  CIGAR~\cite{liu2023cigar} & CVPR 2023 & FCOS & 48.5\\
  HT~\cite{deng2023harmonious} & CVPR 2023 & FCOS & 60.3\\
  \hline
  SFA~\cite{wang2021exploring} & ACM MM 2021 & Def DETR & 46.7\\
  DA-DETR~\cite{zhang2023detr} & CVPR 2023 & Def DETR & 48.9 \\
  \hline
  Source Only & - & RetinaNet & 44.7 \\
  DA-RetinaNet~\cite{pasqualino2021unsupervised} & CVIU 2021 & RetinaNet & 53.6 \\
  Ours & - & RetinaNet & \textbf{61.7}\\
  \hline
\end{tabular}
\end{table}

\subsection{Implemental Details}
We take RetinaNet~\cite{lin2017focal} as the base detector for experiments following~\cite{pasqualino2021unsupervised,tian2023domain}. The  VGG16~\cite{simonyan2015very} pre-trained on ImageNet~\cite{russakovsky2015imagenet} is adopted as the backbone. The student detector is optimized using SGD with a learning rate of 0.01 and a momentum of 0.9, and the weight decay is set to 0.0001. Both the unlabeled data weight $\omega$ and the sensitivity coefficient $\gamma$ are set to 2. Following~\cite{zhou2022dense}, we adopt an alternative ratio threshold $k\%$ for $\tau_p$ specifying $k = 1$, i.e., we select the anchor-level predictions with top one percent scores in an image as the candidates for pseudo labeling, and the rest are suppressed to background. We have a low uncertainty threshold $\tau_u$ of 0.12. The teacher detector is updated via EMA with a weight smooth of 0.9995. We pre-train the student detector for 10k iterations with labeled source data to initialize the teacher detector, and then continue our training for 30k iterations using data from both the source and target domains. We follow the same data prepossessing and augmentation pipeline described in~\cite{xu2021end}. All the training experiments are conducted on 2 RTX 3090 GPUs with 8 images (4 labeled and 4 unlabeled images) per GPU.

\subsection{Comparison with State-of-the-arts}
Following~\cite{hsu2020every, chen2022learning, he2023bidirectional}, we compare with state-of-the-art methods in three different scenes of domain shifts, and report the results with a IoU threshold of 0.5.

\textbf{Adaptation in inverse weather.} 
In this scenario, we evaluate the effectiveness of adaptation methods between similar domains. We take the training set of Cityscapes and Foggy Cityscapes in our experiment and report the results on the validation set of Foggy Cityscapes where the domain shift is caused by weather conditions. As shown in Table~\ref{tab:ctf}, the proposed method outperforms all the listed state-of-the-art results using two-stage, one-stage, and set prediction detection pipelines by 5.0\%, 5.5\%, and 4.7\%, respectively. Furthermore, we can observe that our method achieves the highest AP on five categories, i.e., person, rider, car, motor, and bicycle. For the averaged performance, our MGCAMT outperforms DA-RetinaNet and KTNet by 10.3\% and 10.0\%, which is significant as the two models mentioned also use RenitaNet as base detector.

\begin{table}[t]
\centering
\caption{Experimental results from  Sim10K to Cityscapes. We use VGG16 as the backbone network. Average precision (\%) is evaluated on target images.} \label{tab:stc}
\renewcommand{\arraystretch}{1}
\setlength\tabcolsep{2pt}
\setlength\arrayrulewidth{0.2mm}
\begin{tabular}{c|c|c|c}
  \hline
  Method & Venue & Detector & AP of car\\
  \hline
  SCDA~\cite{zhu2019adapting} & CVPR 2019 & Faster R-CNN & 43.0\\
  HTCN~\cite{chen2020harmonizing} & CVPR 2020 & Faster R-CNN & 42.5 \\
  SIR~\cite{chen2021sequential} & TIP 2021 & Faster R-CNN & 46.0 \\
  D\_adapt~\cite{jiang2021decoupled} & ICLR 2022 & Faster R-CNN & 50.3\\
  TDD\cite{he2022cross} & CVPR 2022 & Faster R-CNN & 53.4 \\
  PT\cite{chen2022learning} & ICML 2022 & Faster R-CNN & 55.1 \\
  \hline
  EPM~\cite{hsu2020every} & ECCV 2020 & FCOS & 49.0\\
  SSAL~\cite{munir2021ssal} & NeurIPS 2021 & FCOS & 51.8\\
  SIGMA~\cite{li2022sigma} & CVPR 2022 & FCOS & 53.7\\
  OADA~\cite{yoo2022unsupervised} & ECCV 2022 & FCOS & 59.2\\
  CIGAR~\cite{liu2023cigar} & CVPR 2023 & FCOS & 58.5\\
  HT~\cite{deng2023harmonious} & CVPR 2023 & FCOS & 65.5\\
  \hline
  SFA~\cite{wang2021exploring} & ACM MM 2021 & Def DETR & 52.6\\
  AQT~\cite{huang2022aqt} & IJCAI 2022 & Def DETR & 53.4\\
  O$^2$Net~\cite{gong2022improving} & ACM MM 2022 & Def DETR & 54.1\\
  DA-DETR~\cite{zhang2023detr} & CVPR 2023 & Def DETR & 54.7 \\
  BiADT~\cite{he2023bidirectional} & ICCV 2023 & Def DETR & 56.6\\
  MRT~\cite{zhao2023masked} & ICCV 2023 & Def DETR & 62.0 \\
  MTM~\cite{weng2023mean} & AAAI 2024 & Def DETR & 58.1\\
  \hline
  Source Only & - & RetinaNet & 45.5 \\
  DA-RetinaNet~\cite{pasqualino2021unsupervised} & CVIU 2021 & RetinaNet & 58.0 \\
  Ours & - & RetinaNet & \textbf{67.5}\\
  \hline
\end{tabular}
\end{table}

\begin{table}[t]
\begin{center}
\caption{Ablation experiments results from Cityscapes to Foggy Cityscapes. We use VGG16 as the backbone network. Mean average precision (\%) is evaluated on target images. CCA, TCA, and FCA denote EDL Guided Classification Confidence Alignment, Interactive Remapping based Task Confidence Alignment, and Focusing Confidence Alignment, respectively.} \label{tab:ablation}
\vspace{-0.1cm}
\resizebox*{7.65cm}{!}{
\begin{tabular}{>{\centering\arraybackslash}p{1.5cm}|*{3}{>{\centering\arraybackslash}p{0.9cm}}|>{\centering\arraybackslash}p{0.8cm}}
  \hline
  Method & \makecell{CCA} & \makecell{TCA} & \makecell{FCA} & \makecell{ mAP }\\
  \hline
  Source Only &   &   &   & 37.4\\
  \hline
    &   &   & \checkmark & 48.8 \\
    & \checkmark &   & \checkmark & 52.4\\
  Proposed &   & \checkmark &   & 45.5 \\ 
    &   & \checkmark & \checkmark & 54.9\\
    & \checkmark & \checkmark & \checkmark & 55.9\\
  \hline
\end{tabular}
}
\end{center}
\end{table}

\begin{figure*}[t]
\centering
\centerline{\includegraphics[width=0.95\textwidth,height=0.635\textheight]{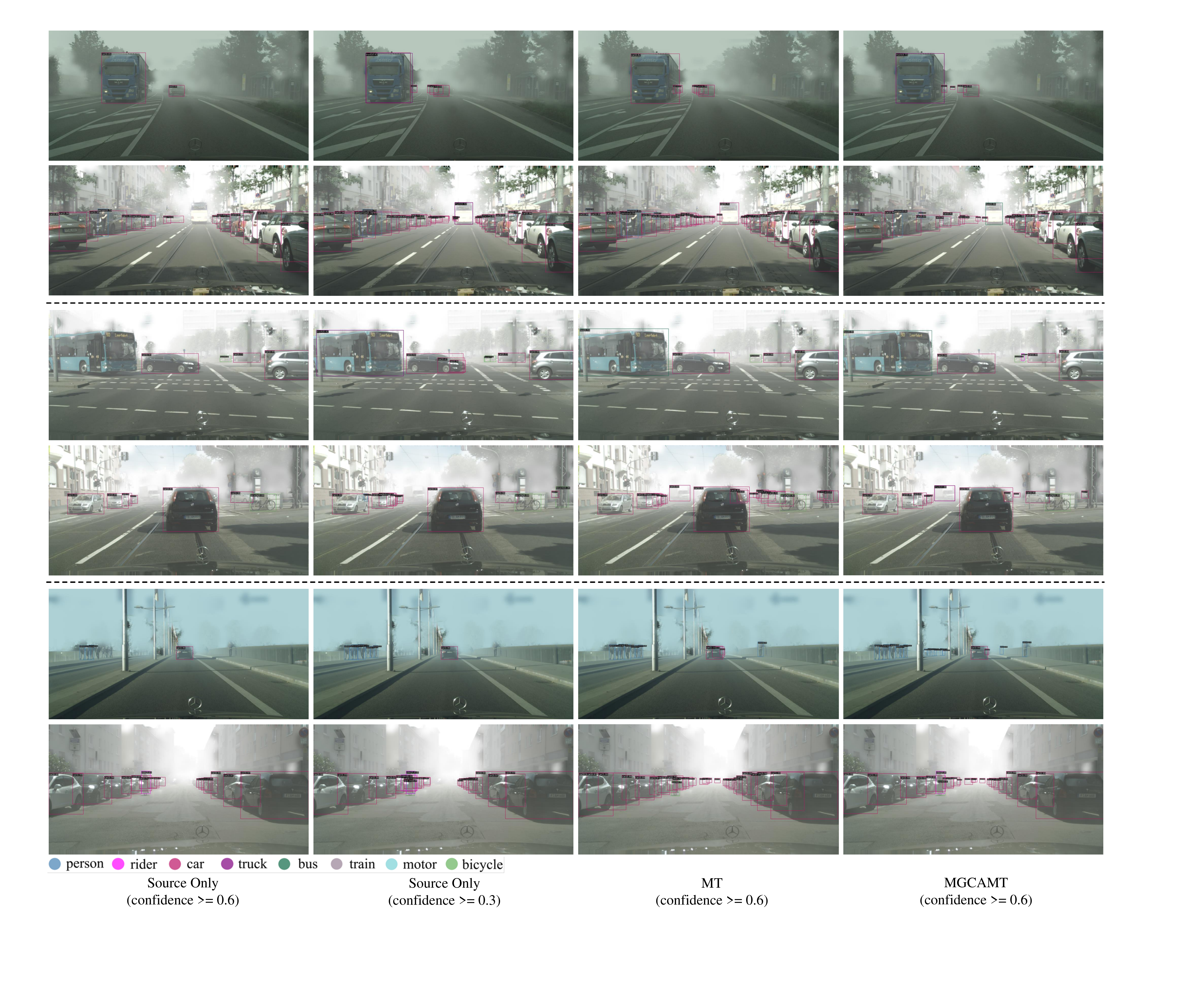}}
\vspace{0.1cm}
\caption{Qualitative results. We compare the detection results of Source Only, MT and MGCAMT with confidence greater than 0.6 on Foggy Cityscapes. Besides, we lower the confidence threshold for Source Only to obtain more comprehensive predictions. MGCAMT fixes misclassification (top, the truck and the bus), inaccurate localization (middle, the bus and the car), false negative and false positive (down, the dense small person and car). (Best viewed in color and zooming in.)}
\label{fig:box}
\end{figure*}

\textbf{Adaptation to a different camera modality.}
In this experiment, we specialize in cross-camera migration. Because of different intrinsic and extrinsic camera parameters, such as resolution, distortion, installation height, and orientation, the objects of interest captured by different cameras are different from each other. We use Cityscapes as the source domain dataset, with BDD100k containing distinct attributes as the unlabeled target domain dataset. Adaptation results are recorded in Table~\ref{tab:ctb}. The proposed MGCAMT outperforms all the recent works by 9.9\% mAP over two-stage adaptive detector, 4.6\% mAP over one-stage method, and 7.5\% mAP over set prediction detection approach. In most categories, we achieve the highest performance.

We conducted more comprehensive experiment adapting KITTI to Cityscapes. The KITTI dataset serves as the source domain, and the Cityscapes dataset as the target domain, with its validation set used for evaluation. Following the literature, the detection accuracy is reported on car category. According to Table~\ref{tab:ktc}, the proposed method achieves an improvement of 1.4\% over the best detector reported on this dataset.  

\textbf{Adaptation from synthetic to real images.}
Synthetic images provide an alternative solution to alleviate the data collection and annotation problems. However, there is a distribution gap between synthetic data and real data. In this section, we study cross-domain car instance detection between these two types of data. During training, Sim10k is employed as the source domain, and the training set of Cityscapes as the target domain. The test is performed on the validation images in Cityscapes. As shown in Table~\ref{tab:stc}, our MGCAMT achieves 67.5 mAP, which exceeds all the other works by 2.0\%.

\subsection{Ablation Studies}
In order to evaluate the effect of each proposed component of our method, we conduct an ablation study. In this subsection, all the studies use Cityscapes and Foggy Cityscapes datasets, as summarized in Table~\ref{tab:ablation}. 

We observe that FCA significantly improves the baseline from 37.4\% mAP to 48.8\% mAP indicating that Focusing Confidence Alignment fully exploits the context information of pseudo labels in the target domain. When we further integrate CCA and TCA separately, there are individual gains of 3.6\% and 6.1\%, respectively. The results demonstrate that these two methods of confidence alignment effectively improve the quality of pseudo labels. Furthermore, as the localization quality improves, the model can obtain complete object information to learn more distinctive and robust features, which is reflected in TCA showing a relatively good performance enhancing. In addition, TCA boosts the performance of the baseline to 45.5\% mAP when employing only the source domain. It shows that the component learns better intrinsic consistency between classification and localization, enhancing the generalization of the model. We achieve the best performance of 55.9\% mAP with all three components, evidencing that our proposed approach maximizes the advantages of each module. In addition, the ablation experiments reveal that these three confidence alignment methods can promote each other and work better cooperatively.



\begin{figure*}[t]
\centering
\centerline{\includegraphics[width=0.945\textwidth]{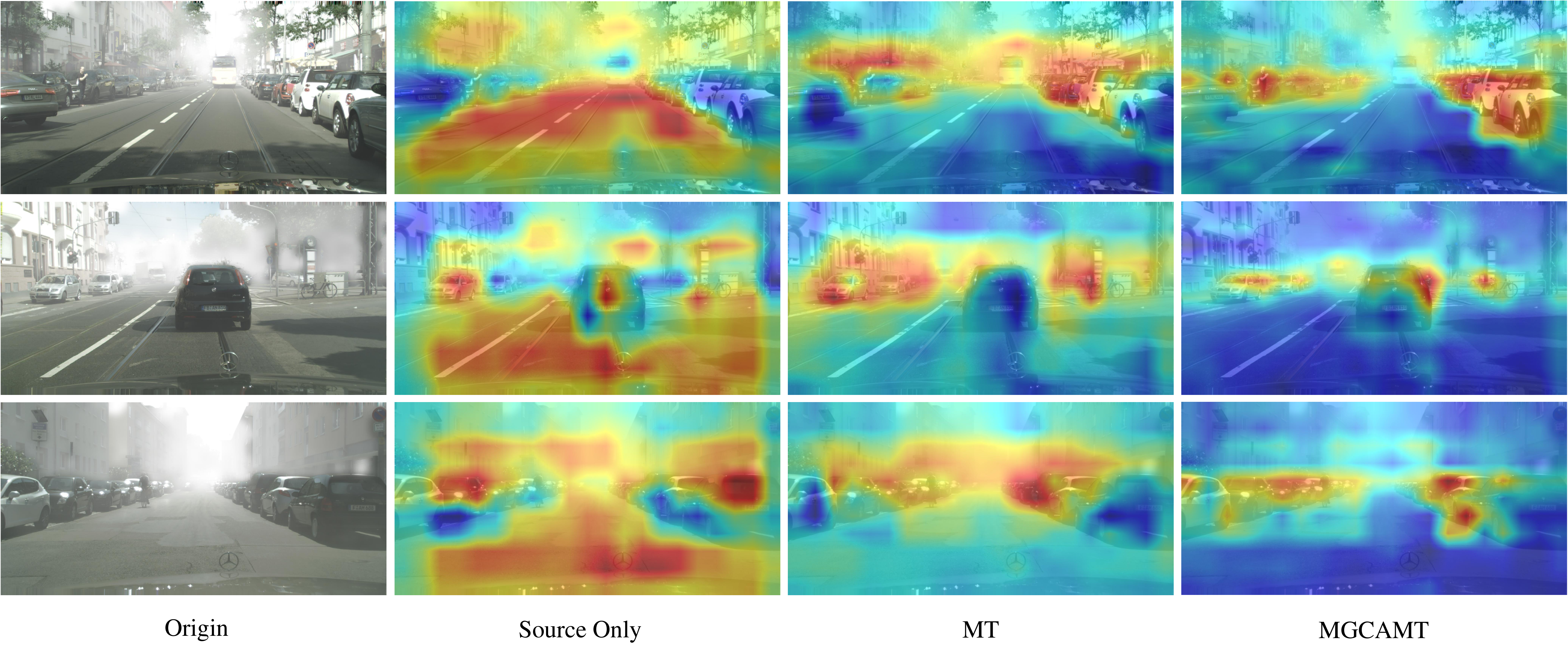}}
\caption{Illustration of image level features for the adaptation from Cityscapes to Foggy Cityscapes, in which Source Only, MT and MGCAMT method are evaluated. We present the origin image in the left. (Best viewed in color and zooming in.)}
\label{fig:heat map}
\end{figure*}

\subsection{Mitigating the Confidence Misalignment}
In this section, we report the mitigating results of confidence misalignment to further validate the effectiveness of our approach. Results on Foggy Cityscapes are presented.

Fig.~\ref{fig:confidence} depicts the classification precision of the predictions in the target domain, with the confidence set to high levels. It can be observed that the proposed MGCAMT exhibits lower error rate for category at higher confidence level, indicating alignment between the classification confidence and actual accuracy. Furthermore, the classification precision improves as the confidence level increases, reaching an accuracy of 93.6\% when the confidence level exceeds 0.8. This further illustrates that our approach have achieved a reasonable self-assurance.

We visualize the confidence of the maximum category and its maximum IoU with the Ground Truth boxes in the corresponding class in Fig.~\ref{fig:task}. Compared to Mean Teacher (MT), our confidence-IoU distribution is concentrated in the upper right corner and along the diagonal. This demonstrates the superiority of the devised approach in obtaining high task consistency and category confidence.

In Fig.~\ref{fig:learning_iteration}, we plot the mean average precision curve of FCA on the target domain. From Fig.~\ref{fig:learning_iteration}, we can see that the detector exhibits a steady improvement under pseudo supervision rather than negative feedback. It is evident that our method effectively alleviates the issue of confidence misfocusing and fully exploits the rich information in the pseudo labels. Through the qualitative detection results in~\ref{fig:box}, we are able to place greater emphasis on the overall objects of interest, further showcasing our superior performance on mutual learning.

\renewcommand{\arraystretch}{1.4}
\begin{table}[t]
\begin{center}
\caption{The EDL hyperparameter of $\tau$ analysis on the adaptation from Cityscapes to Foggy Cityscapes. We use VGG16 as the backbone. Mean average precision (\%) is evaluated on target images.} \label{tab:hyperparameter}
\vspace{-0.1cm}
\resizebox*{7.8cm}{!}{
\begin{tabular}{*{1}{>{\centering\arraybackslash}p{0.9cm}}|*{5}{>{\centering\arraybackslash}p{0.8cm}}}
  \hline
  $\tau$ & 0.13 & 0.12 & 0.11 & 0.10 & 0.09\\
  \hline
  mAP & 55.1 & 55.9 & 55.8 & 55.8 & 55.3\\
  \hline
\end{tabular}
}
\end{center}
\end{table}

\begin{figure*}[t]
\centering
\centerline{\includegraphics[width=0.90\textwidth]{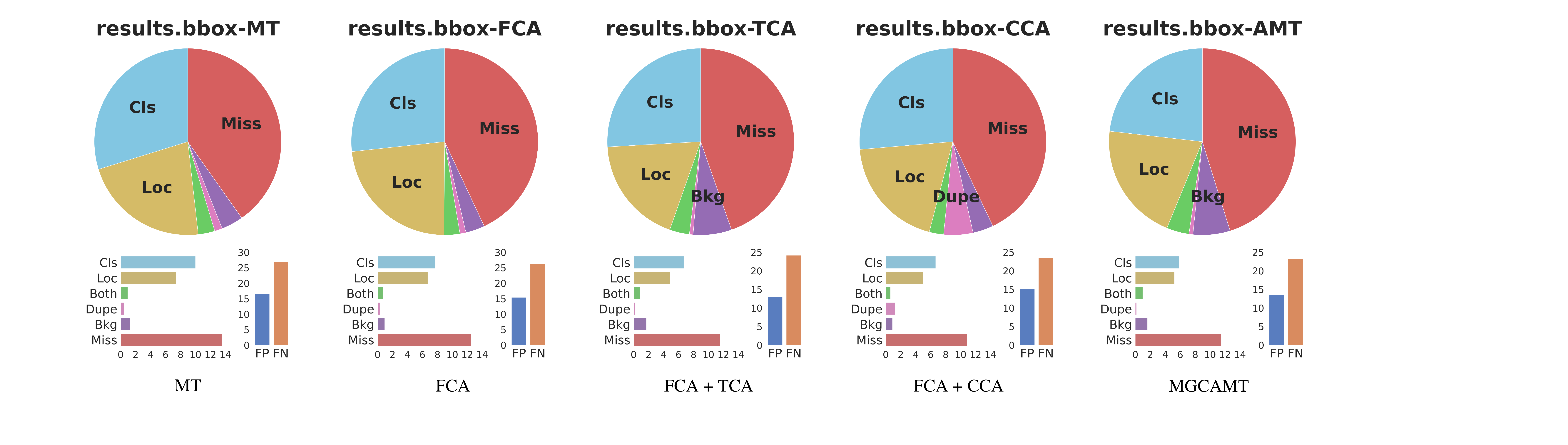}}
\caption{Error analysis on Foggy Cityscapes of MT, FCA, FCA+TCA, FCA+CCA and MGCAMT using TIDE~\cite{bolya2020tide} toolbox. (Best viewed in color and zooming in.)}
\label{fig:error analysis}
\end{figure*}

\subsection{Influence of EDL Hyperparameter}
Our framework introduces new threshold hyperparameters $\tau_p$ and $\tau_u$. Following~\cite{zhou2022dense}, we adopt an alternative ratio threshold $k\%$ for $\tau_p$ specifying $k = 1$ to preliminarily confirm the pseudo labels. As shown in ~\ref{tab:ablation}, we achieved an excellent performance of 54.9 mAP on Foggy Cityscapes. Once the confidence threshold is chosen, we adjust the threshold of EDL uncertainty to filter out the pseudo labels. Table~\ref{tab:hyperparameter} shows the influence of EDL hyperparameter of our method. From Table~\ref{tab:hyperparameter}, we can see that using $0.09 < \tau_u < 0.13$ contributes to performance improvement and it achieves optimal result of 55.9 mAP when $\tau_u$ is set to 0.12. We claim that when $\tau_u$ reaches its upper bound, it can eliminate some overconfident pseudo labels to provide better guidance for the student detector in learning the target domain. Within a certain range, the detector will derive benefits from the pseudo supervision signals. As $\tau_u$ decreases further, more pseudo labels are filtered out, leading to a significant reduction in the informative supervision information they contain. Consequently, this results in predictable performance degradation in mutual learning. It is worth noting that the proposed method requires little hyperparameter tuning: we apply the same threshold from Cityscapes to Foggy Cityscapes for the adaptation from Cityscapes to BDD100K, with a slight adjustment to 0.16 for other scenarios (from KITTI to Cityscapes and from Sim10K to Cityscapes).

\subsection{Qualitative Analysis}
\textbf{Detection results.} We present the detection results in the adverse weather conditions scenario in Fig.~\ref{fig:box}. As shown in the figure, we observe that MGCAMT eliminates some classification errors, e.g., the truck and the bus in the top, comparing to Source Only and MT. Moreover, as shown in the middle row, MGCAMT is able to better locate the objects, like the bus and the car. In addition, MGCAMT can detect the dense small targets missed by Source Only and MT, as well as avoid the occurrence of redundant boxes shown in the bottle of the figure. Consequently, we can confirm that our proposed approach works successfully for cross domain object detection.

By analyzing the confidence of the predictions, we can discover that the detector achieves reasonable self-assurance for the objects (see the top of the figure). Meanwhile, the classification confidence is consistent with the localization quality illustrated in the middle row. Moreover, when we encounter complex cross domain scenarios, the detector can focus on holistic information of the image and detect the scattered targets resoundingly. The results of the confidence alignment mentioned above demonstrate the effectiveness of our method.

\textbf{Features Visualization.} To analyze the properties of the features learned by this model, we show image level visualization in Fig.~\ref{fig:heat map}. Compared to Source Only and MT, MGCAMT focuses on objects that are distributed throughout the image instead of paying attention to individual subsets, while attenuating the distraction of some uncorrelated things. Therefore, it demonstrates that our method successfully extracts the features of interest for the given domains as expected.

\textbf{Detection errors.} A detection error analysis using TIDE~\cite{bolya2020tide} toolbox is reported in Fig.~\ref{fig:error analysis}. From the experimental results, we have the following observations. Compared with vanilla Mean Teacher results, our FCA method can reduce the classification error (Cls), localization error (Loc), and missed GT error (Miss) to some extent, certifying the effectiveness in domain adaptation. It also demonstrates that a good imagery context perception contributes to high quality predictions of category and instance profiles. The clear reduction in primary errors e.g. Cls,  Loc, Miss, and false negative (FN) by the proposed TCA and CCA approach further enhance the domain transferability of the model. Furthermore, it is shown that both TCA and CCA have a corrective influence on classification and regression errors, suggesting that the accurate category attributes and complete appearance information can positively reinforce each other. The improvement in Miss and FN confirms that effective predictions of category and instance localization are conducive to imagery context perception. There is a significant decline in the above errors using MGCAMT,  clearly verifying the efficacy of the proposed adaptation framework. 

\begin{figure}[t]
    \begin{minipage}[t]{0.5\linewidth}
        \centering
        \includegraphics[width=\textwidth]{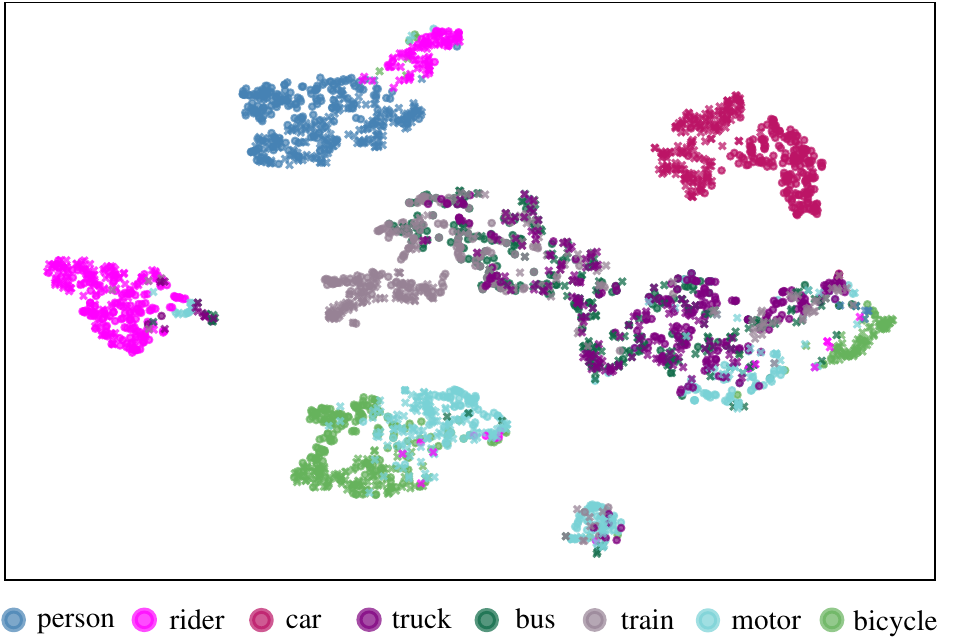}
        \centerline{\fontsize{8}{16} \selectfont MT}
    \end{minipage}%
    \begin{minipage}[t]{0.5\linewidth}
        \centering
        \includegraphics[width=\textwidth]{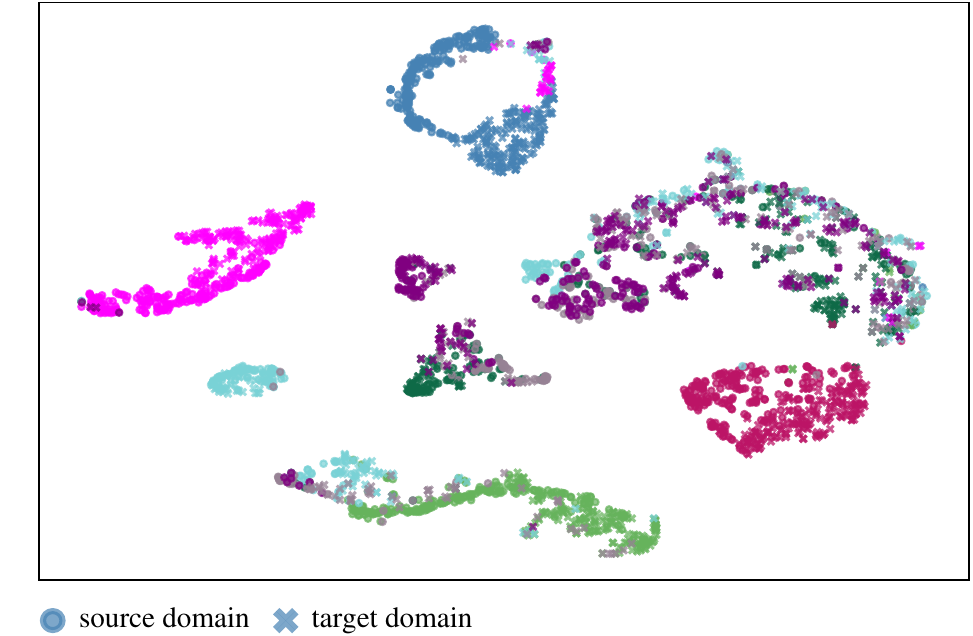}
        \centerline{\fontsize{8}{16} \selectfont MGCAMT}
    \end{minipage}
    \caption{t-SNE visualization of instance level features on source and target domain from Cityscapes to Foggy Cityscapes. The circles are sampled from the source domain, while these cross shapes are from the target domain. We equally sample 200 feature vectors for each class in both domains. (Best viewed in color and zooming in.)}
\label{fig:tsne}
\end{figure}

\textbf{t-SNE visualization.} In Fig.~\ref{fig:tsne}, we utilize t-SNE~\cite{van2008visualizing} to visualize the instance feature distribution of source and target domain. For each class, we equally sample 200 instance feature vectors in both domains. From the experimental results, we observe that the Mean Teacher method effectively aligns the features across different domains. However, it is noted that the features in similar classes, such as bicycle, rider, and person tend to get disentangled into several confused sub-clusters. In addition, categories with a limited number of instances, such as truck, bus, and train do not demonstrate distinct decision boundaries. In comparison to the Mean Teacher method, MGCAMT exhibits better category aggregation and the similar classes are more separable. This visually verifies that the proposed method boosts feature alignment which plays a critical role in achieving high quality object detection.

\section{Conclusion}
In this study, we claim that confidence misalignment including category-level overconfidence, instance-level task confidence inconsistency, and image-level confidence misfocusing, leading to the injection of noise in the training process, will bring sub-optimal detection performance on the target domain. We develop a novel general framework termed Multi-Granularity Confidence Alignment Mean Teacher (MGCAMT) for cross domain object detection. From the perspective of alleviating confidence misalignment across category-, instance-, and image-levels simultaneously, we obtain high quality pseudo supervision to promote teacher-student mutual learning. Within the proposed framework, the Classification Confidence Alignment (CCA) module is devised by introducing Evidential Deep Learning (EDL) to estimate the uncertainty of categories for filtering out the category incorrect ones. Furthermore, we design a Task Confidence Alignment (TCA) module to adjust each classification feature to adaptively locate the optimal feature for the regression task. We also propose a approach named Focusing Confidence Alignment (FCA) to adopt the original outputs from the Mean Teacher network for supervised learning without label assignment. We couple these three confidence alignment methods in a mutually beneficial manner. Comprehensive experiments on widely-used scenarios demonstrate superior performance compared to state-of-the-art methods. Importantly, our study shows that the confidence alignment approach across multiple granularities, which enhances the quality of the pseudo labels, plays a vital role in improving teacher-student learning for cross domain object detection.

\bibliographystyle{IEEEtran}
\bibliography{arxiv}
\begin{IEEEbiography}[{\includegraphics[width=1in,height=1.25in,clip,keepaspectratio]{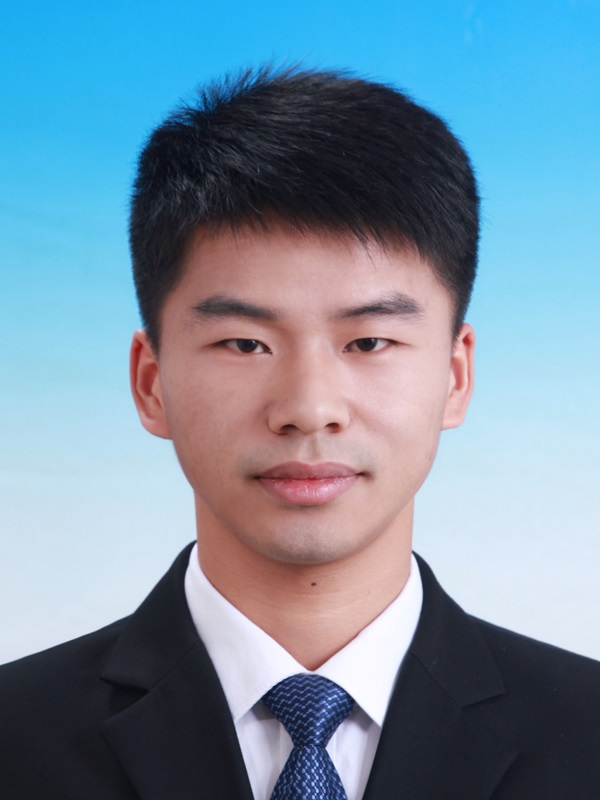}}]{Jiangming Chen}
received the B.S. degree from Xiangtan University, China, in 2011, and the M.S. degree in computer graphics from Beijing University of Aeronautics and Astronautics, china, in 2014. He is currently pursuing the Ph.D. degree with the College of System Engineering, National University of Defense Technology, China. His research interests include computer vision, computer graphics, deep learning and domain adaptation.
\end{IEEEbiography}

\vspace{-10pt}
\begin{IEEEbiography}[{\includegraphics[width=1in,height=1.25in,clip,keepaspectratio]{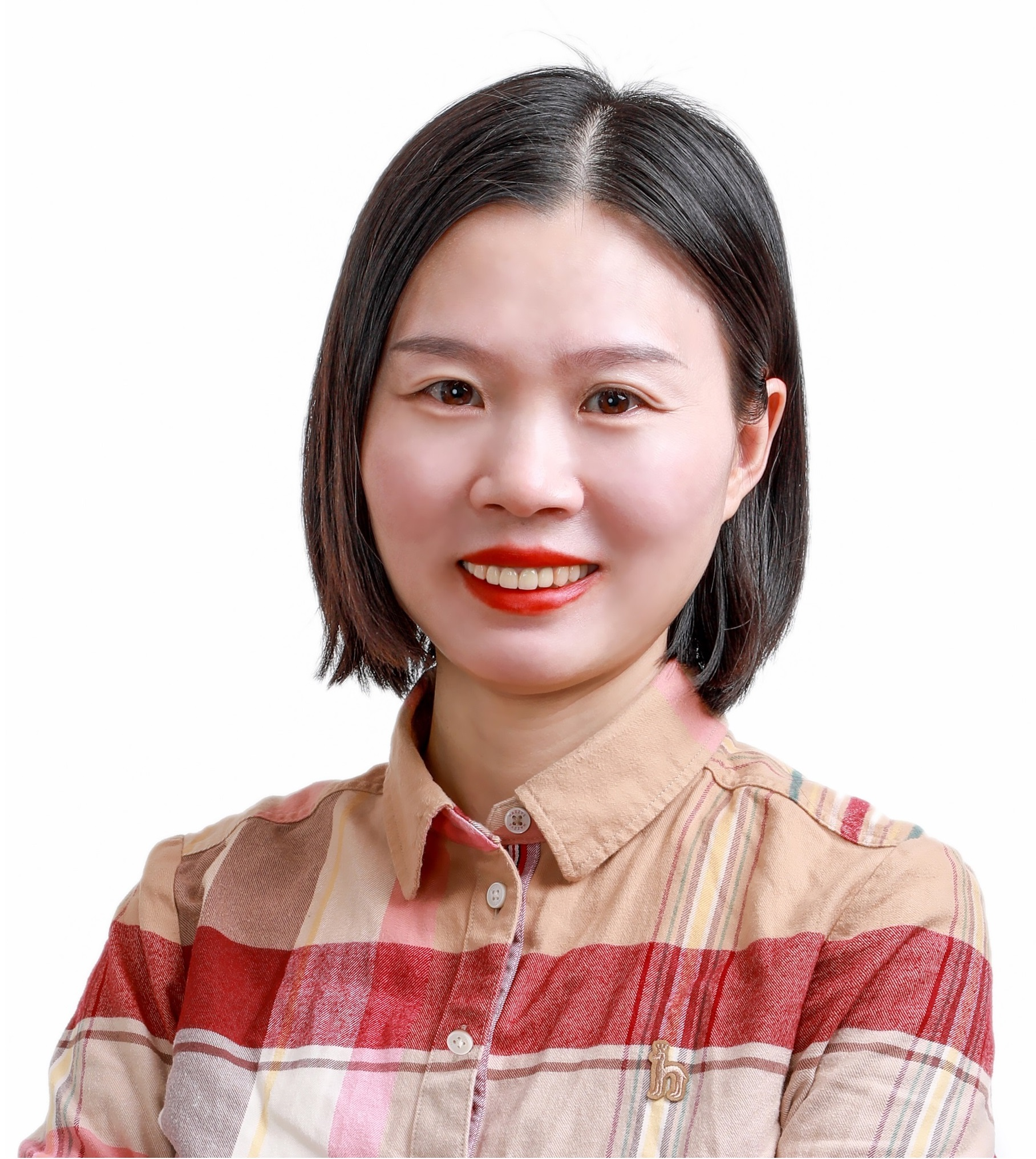}}]{Li Liu}
received the Ph.D. degree in information and communication engineering from the National University of Defense Technology (NUDT), China, in 2012. She is now a full professor with the College of Electronic Science and Technology, NUDT. During her Ph.D. study, she spent more than two years as a visiting student with the University of Waterloo, Canada, from 2008 to 2010. From 2015 to 2016, she spent ten months visiting the Multimedia Laboratory with the Chinese University of Hong Kong. From 2016 to 2018, she worked as a senior researcher with the Machine Vision Group, University of Oulu, Finland. She was a cochair of nine International Workshops with CVPR, ICCV, and ECCV. She served as the leading guest editor for special issues in IEEE TPAMI and IJCV. She is serving as the leading guest editor for IEEE PAMI special issue on “Learning with Fewer Labels in Computer Vision”. Her current research interests include computer vision, pattern recognition and machine learning. Her papers have currently more than 13000 citations according to Google Scholar. She currently serves as associate editor for IEEE TGRS, IEEE TCSVT, and PR.
\end{IEEEbiography}

\begin{IEEEbiography}[{\includegraphics[width=1in,height=1.16in,clip,keepaspectratio]{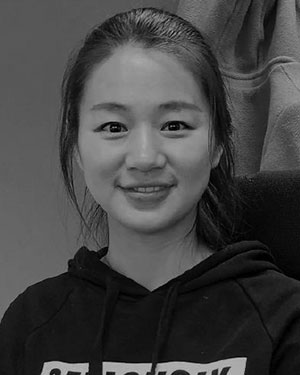}}]{Wanxia Deng} received the B.E. degree in electronic information science and technology from Xiamen University, Xiamen, China, in 2014. She received the M.S. degree and Ph.D. degree in information and communication engineering from the National University of Defense Technology (NUDT), Changsha, China, in 2016 and 2022, respectively. She is currently a  lecturer with the College of Meteorology and Oceanography at NUDT, Changsha, China. Her research interests include domain adaptation, transfer learning, deep learning and AI for science.

\end{IEEEbiography}

\begin{IEEEbiography}[{\includegraphics[width=1in,height=1.2in,clip,keepaspectratio]{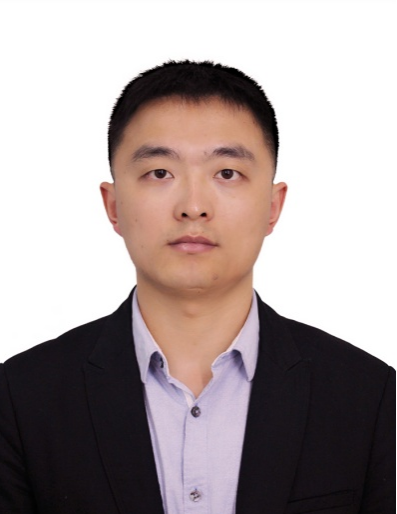}}]{Zhen Liu}
received the Ph.D. degree in Information and Communication Engineering from National University of Defense Technology (NUDT), in 2013. He is currently a professor with the College of Electronic Science and Technology, NUDT. He has been awarded the Excellent Young Scientists Fund on his project titled “Intelligent Countermeasure for Radar Target Recognition” in 2020. His current research interests include radar signal processing, radar electronic countermeasure, compressed sensing, and machine learning.
\end{IEEEbiography}


\begin{IEEEbiography}[{\includegraphics[width=1in,height=1.25in,clip,keepaspectratio]{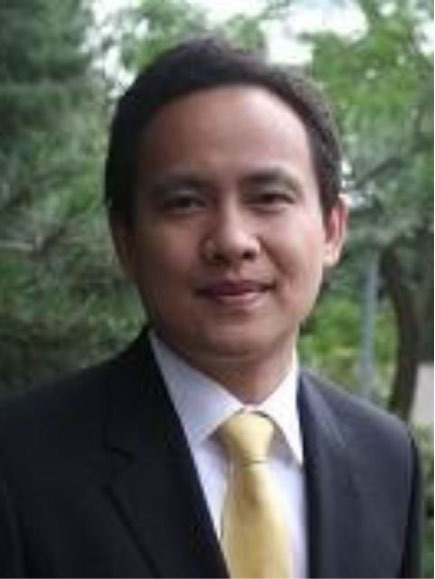}}]{Yu Liu}
received the B.S. degree from Northwestern Polytechnical University, Xi’an, China, in 2005, and the M.S. degree in image processing and the Ph.D. degree in computer graphics from the University of East Anglia, Norwich, U.K., in 2007 and 2011, respectively. He is currently a Professor with the Department of System Engineering, National University of Defense Technology, Changsha, China. His research interests include image/video processing, computer graphics, and machine learning technology.
\end{IEEEbiography}

\begin{IEEEbiography}[{\includegraphics[width=1in,height=1.18in,clip,keepaspectratio]{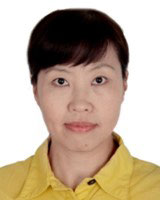}}]{Yingmei Wei}
received the Ph.D. degree in computer science and technology from the National University of Defense Technology (NUDT), China, in 2000. She is currently a Professor with the College of Systems Engineering, NUDT. Her research interests include virtual reality, information visualization, and visual analysis techniques.
\end{IEEEbiography}

\begin{IEEEbiography}[{\includegraphics[width=1in,height=1.30in,clip,keepaspectratio]{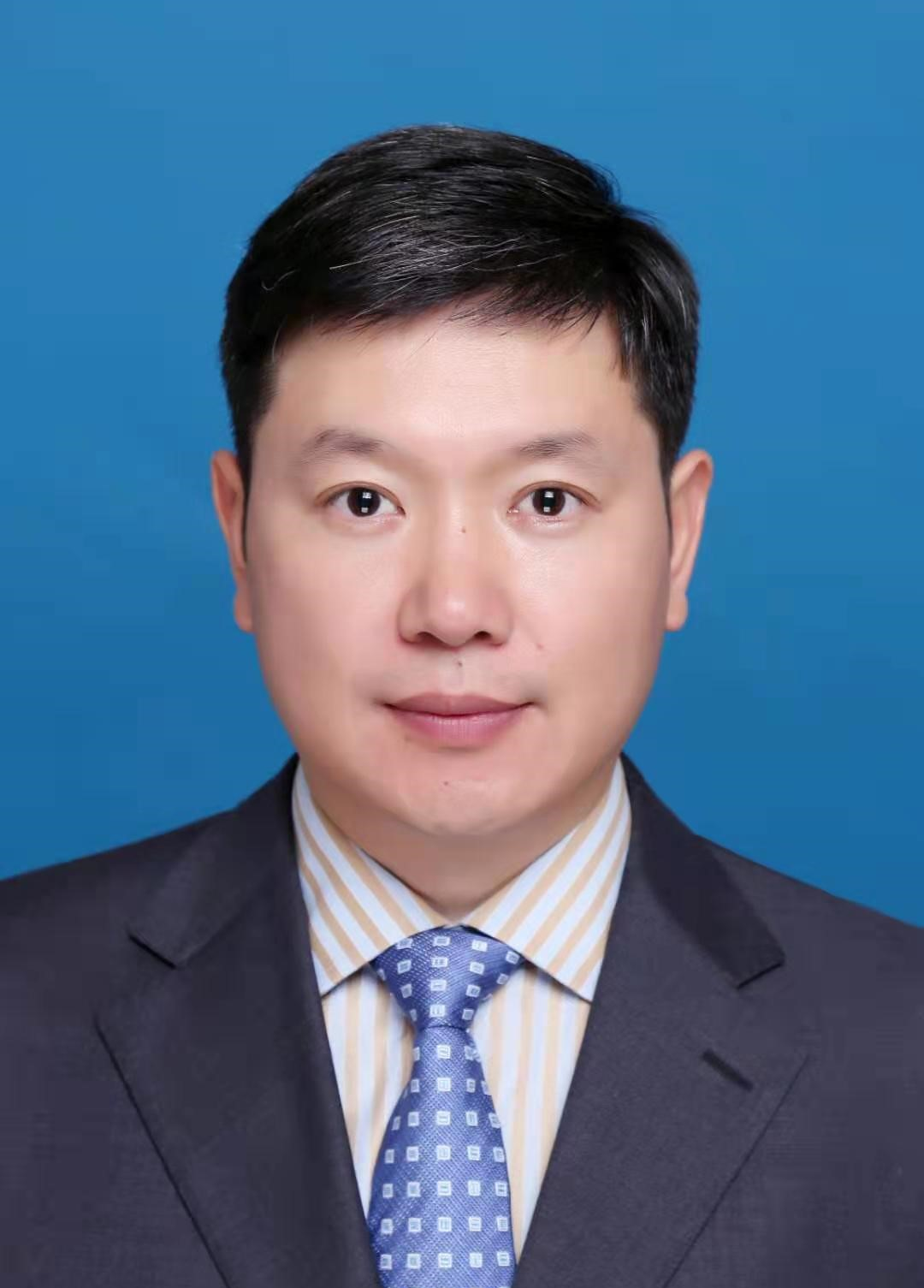}}]{Yongxiang Liu}
received his Ph.D. degree in Information and Communication Engineering from the National University of Defense Technology (NUDT), Changsha, China, in 2004. Currently, He is a Full Professor in the College of Electronic Science and Technology, NUDT. His research interests mainly include remote sensing imagery analysis, radar signal processing, Synthetic Aperture Radar (SAR) object recognition and Inverse SAR (ISAR) imaging, and machine learning.
\end{IEEEbiography}


\vfill

\end{document}